\documentclass
{article}

\PassOptionsToPackage{numbers, compress}{natbib}

\usepackage[final]{neurips_2020}

\usepackage[utf8]{inputenc} 
\usepackage[T1]{fontenc}    
\usepackage{url}            
\usepackage{booktabs}       
\usepackage{amsfonts}       
\usepackage{nicefrac}       
\usepackage{microtype}      
\usepackage{soul,color}

\usepackage{graphicx}
\usepackage{subcaption}
\usepackage{enumitem}

\usepackage{csquotes}
\usepackage{todonotes}
\usepackage{xspace}
\usepackage{amsmath}
\usepackage{amssymb}
\usepackage{float}
\usepackage{wrapfig}
\usepackage{lipsum}
\usepackage{xcolor,colortbl}
\usepackage{fdsymbol}
\usepackage{multirow}
\usepackage{tcolorbox}
\usepackage{rotating}
\usepackage{algorithm,algpseudocode}
\usepackage{pifont}
\usepackage[normalem]{ulem}
\usepackage{makecell}
\usepackage{bbm}
\usepackage{listings}
\usepackage{multicol}
\usepackage{adjustbox}

\usepackage[frozencache,cachedir=.]{minted}

\usepackage{hyperref}

\setminted{style=default}

\definecolor{color1}{HTML}{da6752}
\definecolor{color2}{HTML}{5573a6}
\definecolor{deepblue1}{HTML}{004d5c}
\definecolor{deepmagenta1}{HTML}{5c004b}
\definecolor{green1}{HTML}{0b5400}
\definecolor{orange1}{HTML}{f3905c}
\definecolor{orange}{HTML}{ff7700}
\definecolor{cyan}{HTML}{008b8b}
\definecolor{blue}{HTML}{0000ff}
\definecolor{lb}{HTML}{f0f6ff}
\definecolor{ly}{HTML}{ffffe8}
\definecolor{lg}{HTML}{edfff0}
\definecolor{lr}{HTML}{ffebeb}
\definecolor{lp}{HTML}{f8e8ff}
\definecolor{lgy}{HTML}{f7f7f7}
\definecolor{purple1}{HTML}{9258cc}
\definecolor{blue1}{HTML}{027db5}
\definecolor{cyan1}{HTML}{00a0b5}
\definecolor{pink1}{HTML}{ff7a7a}
\definecolor{red1}{HTML}{8a0000}
\definecolor{blue2}{HTML}{041480}
\definecolor{yellowishgreen}{HTML}{2e8a00}
\definecolor{magenta}{HTML}{9b00a1}
\definecolor{darkgreen}{HTML}{005e19}
\definecolor{darkblue}{HTML}{240394}
\definecolor{darkyellow}{HTML}{826600}
\definecolor{darkred}{HTML}{C00000}
\definecolor{orange2}{HTML}{de7c04}
\definecolor{barplotgreen}{HTML}{78B5A1}

\definecolor{codegreen}{rgb}{0,0.6,0}
\definecolor{codegray}{rgb}{0.5,0.5,0.5}
\definecolor{codepurple}{rgb}{0.58,0,0.82}
\definecolor{backcolour}{rgb}{0.95,0.95,0.92}

\newcommand{\code}[1]{\texttt{#1}}

\newcommand{\bp}{\textcolor{darkred}{\`{}\`{}\`{}}}
\newcommand{\action}[1]{\bp\code{#1}\bp}

\newcommand{\ours}{\texttt{debug-gym}\xspace}

\newcommand{\aider}{Aider\xspace}
\newcommand{\swe}{SWE-bench\xspace}
\newcommand{\night}{Mini-nightmare\xspace}

\newcommand{\agentrw}{\textit{rewrite}\xspace}
\newcommand{\agentpdb}{\textit{debug}\xspace}
\newcommand{\agentdfive}{\textit{debug(5)}\xspace}

\lstdefinestyle{mystyle}{
    backgroundcolor=\color{backcolour},   
    commentstyle=\color{codegreen},
    keywordstyle=\color{magenta},
    numberstyle=\tiny\color{codegray},
    stringstyle=\color{codepurple},
    basicstyle=\ttfamily\footnotesize,
    breakatwhitespace=false,         
    breaklines=true,                 
    captionpos=b,                    
    keepspaces=true,                 
    numbers=left,                    
    numbersep=5pt,                  
    showspaces=false,                
    showstringspaces=false,
    showtabs=false,                  
    tabsize=2
}

\definecolor{dblue}{HTML}{033394}
\definecolor{dgreen}{HTML}{005e19}
\definecolor{dred}{HTML}{8a0000}
\definecolor{dyellow}{HTML}{a37800}

\title{
\ours: A Text-Based Environment for \\ Interactive Debugging}

\author{
Xingdi Yuan$^\spadesuit$\thanks{\:\:\:\:Equal contribution. Full authorship contribution statements are listed in Appendix~\ref{app:contributions}.}\:\: Morgane M Moss$^{\spadesuit\clubsuit}$\:\:
Charbel El Feghali$^\diamondsuit$\\
\textbf{Chinmay Singh$^\heartsuit$\:\: Darya Moldavskaya$^\heartsuit$\:\: Drew MacPhee$^\spadesuit$\:\: Lucas Caccia$^\spadesuit$} \\
\textbf{Matheus Pereira$^\spadesuit$\:\: Minseon Kim$^\spadesuit$\:\:Alessandro Sordoni$^\spadesuit$\:\:Marc-Alexandre C\^{o}t\'{e}$^\spadesuit$\footnotemark[1]} \\
$^\spadesuit$Microsoft Research Montr\'eal \:\:\:\: $^\heartsuit$Microsoft Research NYC \\ 
$^\diamondsuit$McGill University \:\:\:\: $^\clubsuit$Mila, Universit\'e de Montr\'eal\\
\texttt{debug-gym@microsoft.com}}

\begin{document}

\maketitle

\begin{abstract}
Large Language Models (LLMs) are increasingly relied upon for coding tasks, yet in most scenarios it is assumed that all relevant information can be either accessed in context or matches their training data. We posit that LLMs can benefit from the ability to interactively explore a codebase to gather the information relevant to their task. To achieve this, we present a textual environment, namely \ours, for developing LLM-based agents in an interactive coding setting.
Our environment is lightweight and provides a preset of useful tools, such as a Python debugger (\code{pdb}), designed to facilitate an LLM-based agent's interactive debugging.
Beyond coding and debugging tasks, this approach can be generalized to other tasks that would benefit from information-seeking behavior by an LLM agent.
\end{abstract}

\section{Introduction}
\label{sec:intro}


Program synthesis, or code generation, is the task in which a system generates a program based on artifacts that establish semantic and syntactic requirements for the generated code \cite{armando23intro}.
Such artifacts typically consist of natural language descriptions associated with a set of test cases, i.e., a system is required to generate a program that fulfills the natural language descriptions, and can pass the test cases~\citep{chen2021evaluating,hendrycksapps2021,gehring2024rlefgroundingcodellms,jiang2024surveylargelanguagemodels}. 
Large Language Model (LLM)-based systems have shown great promise in code generation tasks, where emergent abilities such as prompting and in-context learning have pushed the boundaries of this research area, from single function code generation \cite{chen2021evaluating, hendrycksapps2021} to more realistic repository-level code generation \cite{zhang2023repocoder, jimenez2023swe, zan2024codes,bairi2024codeplan}. 
The advancement of code generation technology has further empowered AI coding assistants such as GitHub Copilot\footnote{\url{https://github.com/features/copilot}} and Cursor\footnote{\url{https://www.cursor.com/}}, where developers can delegate trivial and repetitive coding tasks, focusing instead on problem solving, collaboration, and creativity.

Recent years have seen a flourishing of code generation systems taking a neuro-symbolic approach \cite{openhands,yang2024sweagent,xia2024agentless}. 
To solve realistic coding tasks, such systems often leverage multiple expert modules designed to tackle different subtasks \cite{arora2024masai}. 
For instance, code and text understanding modules \cite{austin2021program, ahmad2021unified, nam2023ide,zhang2024autocoderover} can be used to convert the repository and related documentations into representations that the code generator module \cite{nijkamp2022codegen,wang2024executable} can better condition on; code-repairing modules can detect and correct potential bugs in the generated code snippets; test generation modules \cite{huang2023agentcoder, alshahwan2024automated,prasad2025unit} can generate necessary test cases to validate the generated code against certain criteria (e.g., functionality, efficiency).
Among these, we concern ourselves in this paper with improving AI-powered code-repairing, a capability that is still not attained \cite{olausson2023self}. 
We believe the elusiveness of code-repairing capability is a bottleneck preventing AI coding systems from producing desirable code, especially when there is an inevitable distribution gap between the system's training data and the user's specific scenarios (i.e., the system cannot generate the correct code in one-shot). 

Most LLM-based code-repairing systems rely on execution feedback \cite{madaan2024self, zhang2023self, jiang2023selfevolve, chen2023teaching}. Given a piece of buggy code, they execute it (e.g., with a Python interpreter) and obtain some error message. Conditioned on this message, the system rewrites the code to fix the bugs. This loop is iterated until the error message is empty, or the agent has exhausted some pre-defined budget (measured in steps or tokens)~\citep{gehring2024rlefgroundingcodellms}. 
While this iterative approach improves repair performance, it might fail when bugs appear in complex real-world software projects, where the error messages can be nested or non-crashing, making them harder to detect and interpret.
\cite{zhong2024ldb} suggest that this may due to LLM-based system's inability to accurately track variable values and predict execution flow.
To address this limitation,~\cite{chen2023teaching} and ~\cite{hu2024leveraging} take inspiration from `rubber duck debugging', a technique where human developers explain code to a rubber duck to clarify their own understanding, or to guide agents to generate \code{print()} functions to help investigate the variable values. 

\begin{figure}[t!]
    \centering
\includegraphics[width=0.95\linewidth]{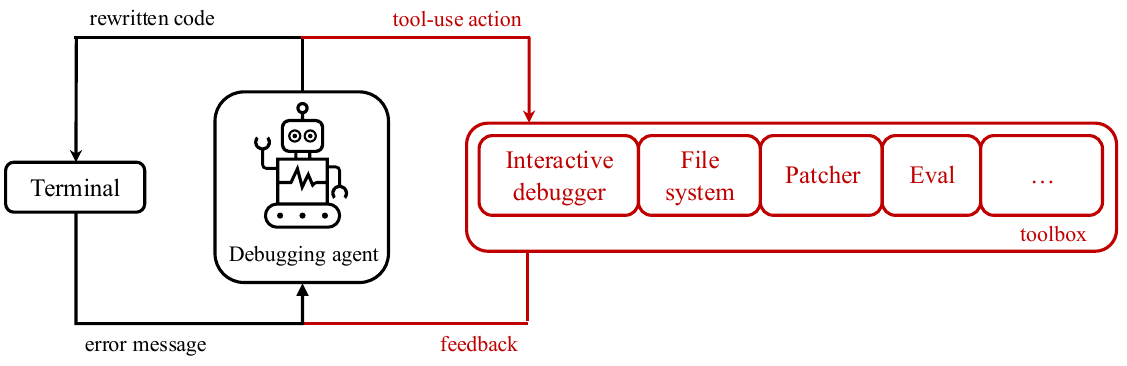}
    \caption{Diagram demonstrating the code-repairing process in outline. In most existing approaches (shown in \textbf{black}), an agent rewrites its code conditioned on error messages obtained from executing the code. \ours equips the agent with additional tools such as \code{pdb} (shown in \textbf{\textcolor{darkred}{red}}), so it can interactively seek necessary information from the semantic space hidden behind the code, and therefore have better code-repairing performance.}
    \label{fig:intro_diagram}
\end{figure}

In addition to talk to a rubber duck friend, or to insert arbitrary numbers of \code{print()} calls into the code, expert developers also rely on interactive debugging tools that are specifically designed to assist in debugging.
In the Python programming language, \code{pdb}\footnote{\url{https://docs.python.org/3/library/pdb.html}} is such a tool.
\code{pdb} allows users to navigate the codebase through breakpoints and other granular stepping functions, 
they can inspect stack frames, list source code chunks of interest, and execute arbitrary Python code in the context of any stack frame.
This enables developers to verify their hypothesis about their code's underlying logic, and thus gain a much more comprehensive understanding of potential bugs.
A natural \textbf{research question} we ask is: 
\begin{center}
\emph{to what degree can LLMs use interactive debugging tools such as pdb?}
\end{center}

To help answer the question, this technical report describes \ours , an interactive coding environment that allows code-repairing agents to access a collection of tools designed to support active information-seeking behavior \cite{gottlieb2013information,bachman2016towards}, such as \code{pdb}.
As depicted in~Figure~\ref{fig:intro_diagram}, \ours expands a debugging agent's action space with a toolbox, which consequently expands the agent's observation space with feedback messages returned from using a tool. 
The toolbox is designed to facilitate debugging: for example, the agent can make use of the Python debugger \code{pdb} to set breakpoints, navigate the code space, print variable values, and even create test functions on the fly. 
At each step, the agent can either decide to interact with a tool to further investigate the code and gather necessary information, or perform a code rewrite if it is confident in doing so.
We believe that interactive debugging (with proper tools) can empower coding agents to tackle real-world software engineering tasks. We also propose that teaching agents to interactively seek information using tools is central to LLM-based agent research in general.

\ours is designed and developed to:
\begin{enumerate}[leftmargin=*]
    \item \textbf{handle repository-level information}: the full repository is available to agents in \ours, allowing them to navigate and modify files.
    \item \textbf{be robust and safe}: to safeguard both the system and the development process, \ours runs code within Docker containers. This isolates the runtime environment, preventing harmful actions while still permitting thorough testing and debugging.
    \item \textbf{be easily extensible}: \ours was conceived with extensibility in mind and provides practitioners with the possibility of easily adding new tools.
    \item \textbf{be text-based}: \ours represents observation information in structured text (e.g., \code{JSON} format) and defines a simple syntax for text actions, making the environment fully compatible with modern LLM-based agents.
\end{enumerate}

\subsection*{Scope of this technical report}
With this technical report, we open-source the \ours environment to the research community to foster future research in this area. 
The goal of this technical report is not to propose any new algorithms, but rather to demonstrate an alternative path to tackle the difficult code-repairing problem and to encourage the community to push towards solving it.

We structure this technical report in four sections:
\begin{itemize}[leftmargin=*]
    \item In Section~\ref{sec:method}, we provide details on the design of \ours. We formally define the interactive debugging problem, then elaborate how the environment, the toolbox, and an agent interact with each other. Finally, we provide instructions on how users could design their own tools that can be used in the \ours framework. This section is meant to serve as a complementary user manual to the \code{README} file in our repository.
    \item In Section~\ref{sec:experiment}, we introduce three LLM-based agents that have minimal architecture design and prompt engineering.
    We provide some preliminary experimental results to help understand these agents' behavior in interactive debugging environments.
    We analyze the experimental results and highlight the challenges faced by LLM-based agents using interactive tools. 
    \item In Section~\ref{sec:future_work}, we discuss the limitation of the current version of \ours, and propose a few promising future directions for both the framework itself and the research it enables.
    \item In Section~\ref{sec:related}, we conduct a literature review on topics related to LLM-based coding and debugging agents, as well as existing language-based interactive environments that facilitate such research.
\end{itemize}

\section{The \ours Environment}
\label{sec:method}

In this section, we first provide a formalization of the interactive debugging process (Section~\ref{subsec:pomdp}), followed by an overview of the system design of the \ours environment (Section~\ref{subsec:system}). 
Then, we describe \ours's pseudo terminal and its safety properties (Section~\ref{subsec:terminal}).
Finally, we introduce \ours's design of toolbox in Section~\ref{subsec:toolbox}, including the preset of tools we provide with \ours (Section~\ref{subsubsec:built-in-tools}), and an instruction on how users can define their own tools in \ours (Section~\ref{subsubsec:new-tool}). 

\subsection{Problem Setup}
\label{subsec:pomdp}

We consider interactive debugging problems where an agent needs to iteratively interact with an environment, use various tools provided by the environment to investigate a buggy codebase, and eventually patch the code to fix the bugs.
We formalize these problems as Partially Observable Markov Decision Processes (POMDPs) \cite{kaelbling1998planning}, defined by a 6-tuple $(S, A, O, \Omega, \mathcal{T}, R)$.
We use $S$ to denote the state space, $A$ to denote the action space, and $O$ the observation space. 
$\Omega: S\rightarrow O$ is the observation function, $\mathcal{T}: S\times A \rightarrow S$ is the transition function, and $R$ is the reward function. 
Each debugging problem (data point) consists a piece of buggy code $c$ that aims to fulfill a natural language problem description $x$, and a set of test cases $\mathcal{U}$.
Let $C$ be the set of valid implementations of $x$ that can pass the test cases $\mathcal{U}$.
An LLM-based agent implements a policy $\pi$ which aims to transform $c$ into $c^* \in C$ through interaction with the environment.




\paragraph{State space $S$:}
\ours uses a modular design to ensure its generality and extensibility (to be detailed in Section~\ref{subsec:system}).
Consequently, the state space $S$, the observation space $O$, and the action space $A$ can also be decomposed in a modular manner. 
Specifically, the state space can be decomposed as $S = S_{\text{env}} \cup S_{\text{tool}}$, where $S_{\text{env}}$ denotes general environment states such as the code in the repository, the directory structure, the set of test cases $\mathcal{U}$, local variables and their values during execution, and optionally ground-truth solutions.
$S_{\text{tool}} = \bigcup\limits_{i=1}^{|\text{tool}|}S^{i}_\text{tool}$ denotes the union of states specific to each tool. 
For example with the \code{pdb} tool, this may be all of the breakpoints currently being activated.

\begin{figure}[t!]
    \centering
\includegraphics[width=0.9\linewidth]{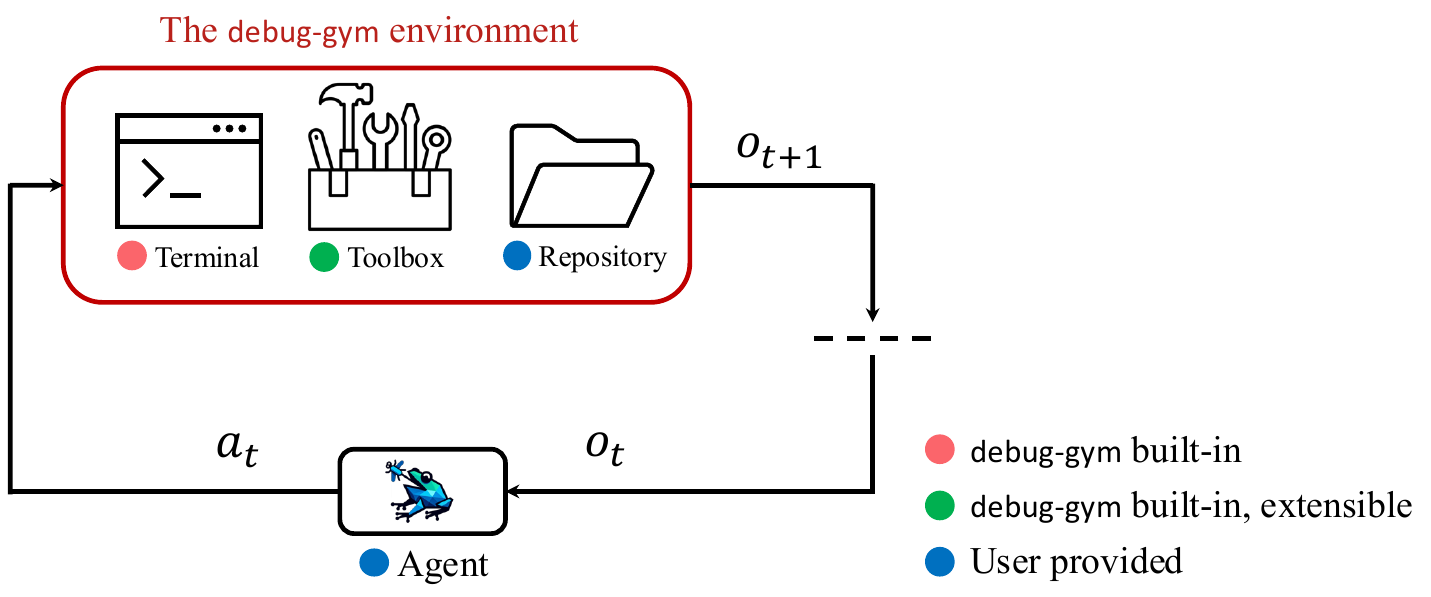}
    \caption{Abstraction of the relationship between components of the \ours interactive debugging system. The environment is defined to accommodate an interactive terminal, an extensible toolbox, as well as a code repository the user aims to investigate. Given the environment, an agent iteratively takes actions in the environment, each one yielding a new observation.}
    \label{fig:abstract-system-overview}
\end{figure}

\paragraph{Observation space $O$:}
The environment is partially observable, an agent has only a limited view of the potentially huge state.
This is due to the fact that it is unrealistic for either human developers or an LLM to include all the available information (e.g., a large repository) into its view / context.
More importantly, the code itself represents an even larger hidden space, including the values of the variables at every step, the control flow the code builds, and the higher level semantics such as algorithms the code implements.
Similar to the state space $S$, the observation space $O$ can also be decomposed into $O = O_{\text{env}} \cup O_{\text{tool}}$, where each individual tool could enrich the agent's observation in some way, i.e., $O_{\text{tool}} = \bigcup\limits_{i=1}^{|\text{tool}|}O^{i}_\text{tool}$.
Through interacting with the environment by utilizing the tools in a proper way (performing information-seeking behavior), the agent should be able to shift its belief from a partial understanding of the codebase towards the true state and thus more easily solve the task.

\paragraph{Action space $A$:}
Based on the needs of solving a particular task, the user can specify a set of tools to be available.
At every step, an agent interacts with \ours via calling a tool from the set of available tools.
Each individual tool has its own syntax, such syntax defines the action space of that tool. 
For example, the \code{rewrite} tool requires a set of arguments such as the file path, the line number to edit, as well as the new content to be written in that position.
The overall action space $A$ is the union of the action spaces over all available tools, i.e., $A = A_{\text{tool}} = \bigcup\limits_{i=1}^{|\text{tool}|}A^{i}_\text{tool}$.
It is worth noting that \ours is a text-based environment, all observations and actions are represented in strings.
If an action string is failed to be parsed, \ours will return an error message of \code{SyntaxError}.

\paragraph{Reward function $R$:}
The reward function $R: S \rightarrow \{0,1\}$ is determined by whether a rewritten code could pass all the test cases $\mathcal{U}$. 
At time step $t$, an agent rewrites the code as $c_t$, it receives a reward:
\begin{equation}
    r_t = R(s_t, a_t) = \begin{cases}
        1 & \text{if } c_t \in C, \\
        0 & \text{otherwise}.
\end{cases}
\end{equation}
In which, $C$ is the set of valid code that can pass all test cases in $\mathcal{U}$.
Note it is often the case where there exists multiple valid solutions and thus comparing $c_t$ with the ground-truth solution $c^*$ (if available) is less applicable.
In case of absence of the test cases, i.e., $\mathcal{U} = \emptyset$, the reward is by default the successfulness of running the code $c_t$ without error message.
An episode is terminated when either an agent has received a positive reward (success), or it has exhausted a pre-defined budget, in the form of the number of steps or the number of consumed tokens (out of budget). 



\subsection{System Design}
\label{subsec:system}


\paragraph{Environment.}
\ours is a Python library, which essentially encompasses the interaction loop between an agent and a repository-specific environment. 
As demonstrated in Figure~\ref{fig:abstract-system-overview}, the environment is an encapsulation of an interactive terminal, a set of tools, a code repository, and optionally a set of test cases to evaluate the correctness of the code repository.
In which, \ours provides the terminal and a preset of tools, the users are required to specify the code repositories they want to investigate, as well as the test cases if applicable. 
To provide a generic interface for developing and training various types of agent systems, the \ours environment adopts an API style similar to the Gym \cite{brockman2016gym} and Gymnasium \cite{towers2024gymnasium}, which have been widely used in the development of modern RL agents. 
Specifically, an \code{env.reset()} function is used to start an episode, whereby the environment returns the initial observation $o_0$; an \code{env.step()} function is used to submit an action $a_t$, the environment subsequently returns the new observation $o_{t+1}$ and the reward $r_t$.
A simplified example of such agent-environment interaction is as follows, here we assume a prompt-based agent:

\noindent\makebox[\linewidth]{\rule{\textwidth}{0.4pt}}
\begin{minted}[linenos,tabsize=4,breaklines,fontsize=\small]{python}
    env = create_env(config)  # Build environment from config
    obs = env.reset()  # Reset/initialize a debugging episode
    reward = 0  # Initial reward is zero
    for step in range(config["max_steps"]):
        # Terminate an episode if success
        if reward > 0:
            break
        # Build LLM prompt
        prompt = agent.build_prompt(obs)
        # Prompting LLM to obtain the next action
        action = agent.llm_call(prompt)
        # Send the action to the environment and get the new observation
        obs, reward = env.step(action)
\end{minted}
\noindent\makebox[\linewidth]{\rule{\textwidth}{0.4pt}}

When building the environment, \ours will copy the user-specified repository into a temporary working directory, which is by default mapped into a Docker container. 
While \ours also supports the option of running the terminal in the user's local machine, which could be faster, we highly recommend users to use Docker as a safeguard which guarantees the actions (e.g., generated by an LLM) do not modify the user's file system in any unexpected way. 
Large code repositories often contain transient files such as temporary files created by IDEs and compilation products, users can include a \code{.debugignore} file in their repository, following the same syntax as \code{.gitignore}, to remove specific files and directories from \ours's scope.
Users can additionally specify read only files or directories using a \code{.debugreadonly} file, in such cases, the specified files and directories are \emph{read only} to an agent through \ours.
This is especially useful for repositories that come with test cases, the read only setting can prevent agents from developing cheating strategies such as to pass the tests by editing or deleting them.


\paragraph{Tools.}
Tools grant the agent a diverse and coherently functional action space with which to facilitate the debugging process. 
For example, the \code{view} tool sets the current working file, and displays its contents in the observation text. 
The \code{rewrite} tool can be used to rewrite all or a portion of the current working file, for example in an attempt to fix a bug in that file.
It can also rewrite a file not in the current view, by specifying a file path.
The \code{pdb} tool interfaces the agent with the full suite of pdb commands that can ordinarily be used in the terminal, allowing the agent to insert breakpoints, inspect local variables, and so on. 
Tools are, in this way, highly modular, and users can introduce their own custom tools to \ours. 
Each tool can have its own syntax depending on the functionality of the tool. 
On top of which, we design a minimal meta-syntax in order for the environment to parse and activate a certain tool at a certain step.
The meta-syntax adopts the triple-backtick delimiter \bp\xspace from Markdown.
For example:\vspace{0.5em}\\
\centerline{\action{\textcolor{darkred}{pdb} cl src/code.py:26}}\vspace{0.5em}\\
calls the \code{pdb} tool and passes:\vspace{0.5em}\\
\centerline{\code{cl src/code.py:26}}\vspace{0.5em}\\\vspace{0.5em}
to the \code{pdb} tool, which aims to delete the breakpoint from line 26 in \code{src/code.py};\\
\centerline{\action{\textcolor{darkred}{rewrite} code/utils.py 4:6 <c>\space\space\space\space\space\space\space\space print('bonjour')</c>}}\vspace{0.5em}\\
calls the \code{rewrite} tool and passes: \vspace{0.5em}\\
\centerline{\code{code/utils.py 4:6 <c>\space\space\space\space\space\space\space\space print('bonjour')</c>}}\vspace{0.5em}\\\vspace{0.5em}
to the \code{rewrite} tool, which replaces the chunk of code between line number 4 and 6 in the file \code{code/utils.py} with the \code{print()} function call, with the indentation ahead (in this case, 8 spaces).
In both examples, meta-syntax are labeled in \textcolor{darkred}{red}, the text in between is the syntax defined by the \code{pdb} and \code{rewrite} tools, respectively.
See Section~\ref{subsubsec:built-in-tools} for a more comprehensive description of tools built-in with \ours, and Section~\ref{subsubsec:new-tool} for more information about adding new custom tools.

\begin{figure}[t!]
    \begin{subfigure}{1\linewidth}
        \caption{A piece of buggy code and the error message returned from a terminal when attempting to run the code.}
        \includegraphics[width=\textwidth]{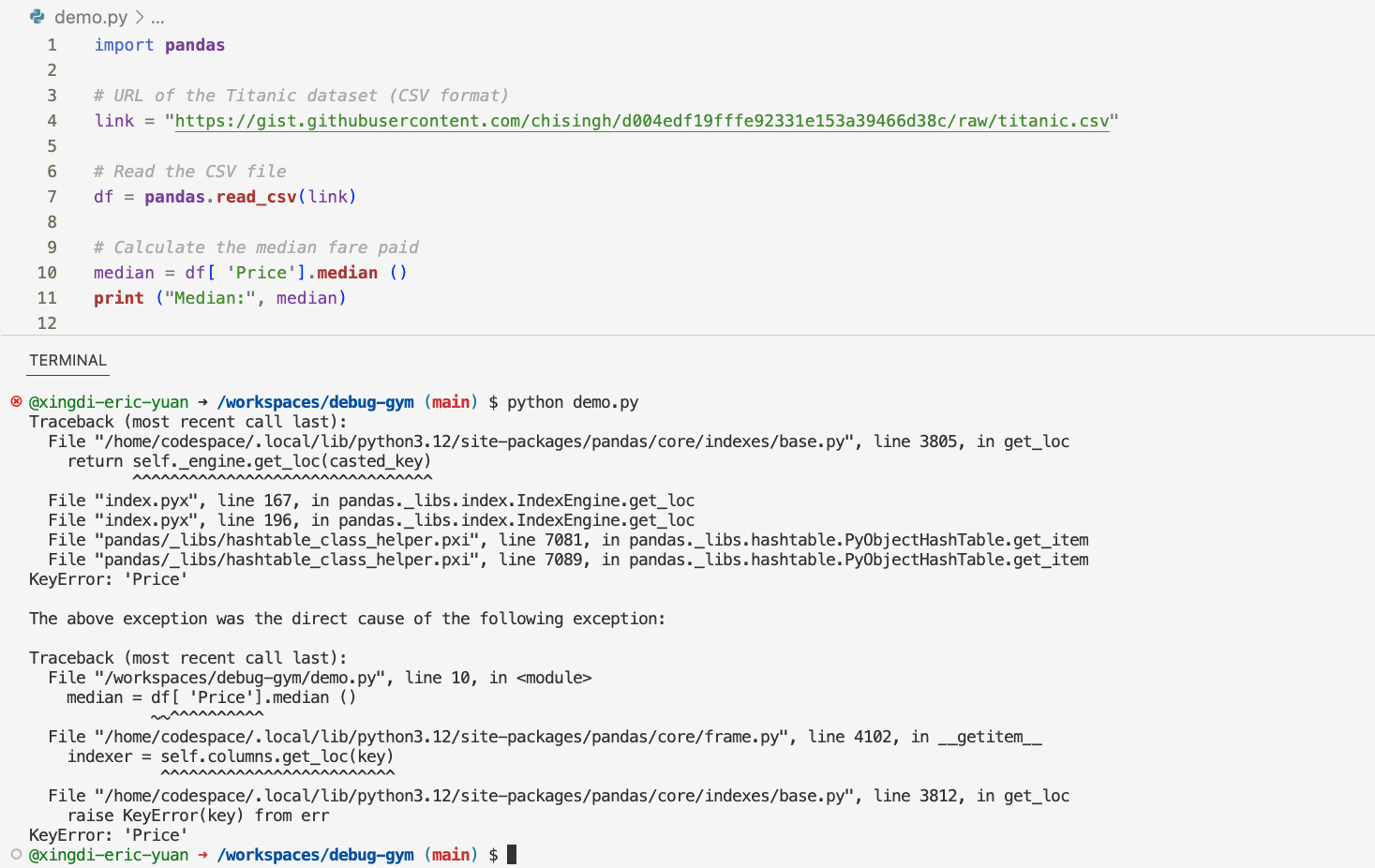}
    \end{subfigure}
    \bigskip
    \begin{subfigure}{1\linewidth}
        \caption{Debugging interactions in \ours, between the environment (in black) and the agent (in \textcolor{darkred}{red}).}
        \includegraphics[width=\textwidth]{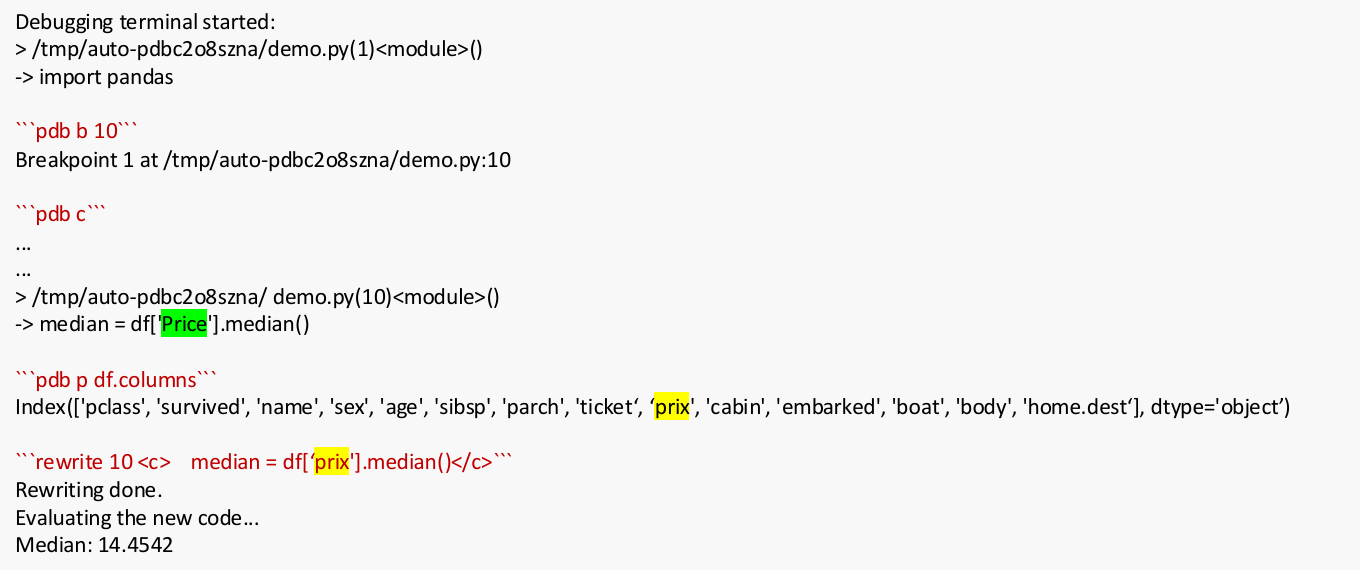}
    \end{subfigure}
\caption{An example demonstrating interactive debugging with \ours. 
    In this example, the agent uses \ours to investigate the columns of the downloaded data frame, and figures out the correct column name is \colorbox{yellow}{prix} instead of \colorbox{green}{price}. 
    This information is unavailable by solely reading the code and the error message, as done by most conventional code-repairing systems.}
\label{fig:example}
\end{figure}

\paragraph{Agent.}
An agent is a system provided by the user, its implementation encompasses the user's design decisions about how the system should interact with \ours to perform the debugging task.
Although the majority of this technical report assumes an LLM-based agent, the implementation of an agent can take many different forms, including rule-based programs, LLM-based chatbots, or even systems that have humans-in-the-loop. 
In any case, at step $t$, the agent takes the observation $o_t$ as input and generates an action $a_t$ as output (as shown in Figure~\ref{fig:abstract-system-overview}). 
Crucially, the agent contains a \code{run()} method, which allows the agent to interface with the environment to solve the task, and therefore the environment is mounted onto the agent instance as an attribute (this is automatically taken care of by the canonical \code{run.py} script provided). 
Customizing this method enables the user to experiment with different ways for the agent to respond to a given observation.
In \ours, we provide a starter kit that includes three minimal LLM-based agent implementations (described in Section~\ref{subsec:example_agents}), serving the purpose of demonstrating \ours's API.
The users can inherit from the provided base agents to design and build their novel and more complex debugging agents.
We provide an example of an agent performs interactive debugging using \ours in Figure~\ref{fig:example}.

\subsection{Terminal}
\label{subsec:terminal}
The terminal provides a way to execute commands from the temporary working directory. For instance, the terminal is used to run the codebase's entrypoint (e.g., \code{pytest tests.py}) to get observations and measure progress, and the debug\_entrypoint (e.g., \code{python -m pdb -m pytest tests.py}) to start a debugging shell session as used by the \code{pdb} tool (see Section~\ref{subsec:toolbox} below for more details about tools).

From a terminal instance, users can execute a standalone command and wait for their output to be returned once done. Users can also start shell sessions that allow interacting with any command line program that requires multiple inputs throughout its execution (e.g., while debugging).

\ours provides two types of terminal: a local one and one running on Docker. While the local terminal is faster to start, it also exposes the user's host filesystem which is unsafe. On the other hand, Docker provides isolation and the ability to easily manage different problems that may require different package requirements. Upon opening a Docker terminal, the temporary working directory is mounted as a Docker volume allowing to make changes to the codebase locally from the host. The docker terminal is configurable, allowing users to specify the base image from which docker container is initialized, environment variables, setup commands to install dependencies, and session commands that are executed whenever a shell session starts (similar to a \code{.bashrc} file).

\subsection{Toolbox}
\label{subsec:toolbox}
The toolbox serves as a registry to keep a record of existing tools.
Tools are callable artifacts (e.g., functions) that are designed to accomplish various sub-tasks an agent may encounter while debugging.
The tools can either be called in a single-turn manner (e.g., issuing a \code{rewrite} action will change certain part of the code repository right away) or in a multi-turn manner (e.g., an agent needs to interact with the \code{pdb} tool for multiple steps in a row to investigate the code).
In both cases, a tool behaves like a mini environment: it takes an action as input, after executing its internal function, it returns a new observation as output.
In which, the internal function can modify the environment state $s_\text{env} \in S_\text{env}$ (e.g., editing the code via \code{rewrite}), or the state specific to the tool $s_\text{tool} \in S_\text{tool}$ (e.g., removing a breakpoint via \code{pdb}).
Note that the internal function does not have to be a Python function, it can essentially be any function $\mathcal{F}: \text{input} \rightarrow \text{output}$, such as an external calculator or a parser.

Each tool is associated with a unique template that is used for the environment to parse, identify, and activate a tool given an action. 
A template is typically the tool's name amid special tokens.
For instance, \action{\textcolor{darkred}{pdb} c} will be parsed by the environment, the \code{pdb} tool will be activated, and the command \code{c} will be fed as the input to the tools. 
To make the tools friendly to LLM-based agents, we also associate each tool with a description string and a list of example strings, both as attributes in the tool class. 
The description is a short paragraph describing the functionality as well as the syntax of the tool.
The example list provides examples and their interpretation in natural language, for example:\vspace{0.5em}\\
\centerline{\action{\textcolor{darkred}{pdb} b 42} \code{to set a breakpoint at line 42 in the current file.}}\vspace{0.5em}
While initializing the environment, the user is required to register the set of tools to be used depending on the task requirement.

\subsubsection{Built-in Tools}
\label{subsubsec:built-in-tools}

We provide a preset of tools in the toolbox:
\begin{enumerate}[leftmargin=*]
    \item \code{eval}: This tool can be triggered to run the entrypoint associated with the repository, which evaluates the current code in the repository. 
    When a set of test cases are provided with the repository, the entrypoint can be invoking the \code{pytest} package (e.g., "\code{python -m pytest -sv .}").
    When the test cases are not provided, the entrypoint can be executing the code (e.g., "\code{python main.py}").
    This tool will return a reward of 1 and terminates the current debugging episode if there is no error detected, and a reward of 0 otherwise.
    By default, this tool is automatically triggered after a \code{rewrite} action, but this can be modified by the user through configuration.
    \item \code{view}: This tool takes a valid file path in the repository as input, it switches and displays the contents of the specified file in the environment's observation, and sets that file as the environment's current working file. 
    Like in many IDEs, we provide line numbers in front of each line of the observing code, this can make the referencing (e.g., when rewriting a line) easier.
    Although there are possible ways to manipulate a file outside the current view (e.g., by specifying a path in \code{rewrite} and \code{pdb} actions), working with a file while observing its content is the most natural way of debugging.
    \item \code{pdb}: This tool is a wrapper that acts as a direct interface between the agent and the full suite of \code{pdb} commands available to the original Python debugger, such as \code{b(reak)}, \code{cl(ear)}, \code{s(tep)}, \code{n(ext)}, \code{c(ontinue)}, and \code{p(rint)}.
    An agent can directly manipulate the breakpoints in the current working file (specified with \code{view}), it can optionally manipulate breakpoints in all visible files in the repository by specifying the file name and its path. 
    Like in many IDEs, once a breakpoint has been set, it is not reset when terminal session ends and restarts (e.g., when issuing a \code{continue} command at the last breakpoint).
    We keep a list of all breakpoints and restore them when there is a terminal session restarts. 
    A breakpoint can be removed upon a \code{c(lear)} command, or when an episode terminates.
    Optionally, the user can deactivate this feature in the configuration file, so the list of breakpoints does not persist when terminal session restarts.
    \item \code{rewrite}: This tool aims to enable an agent to directly modify the code in a file in the repository. 
    In comparison to typical code rewriting strategies that rewrite the entire file (because many widely-used datasets target relatively short function-level code generating \cite{chen2021evaluating,gauthier2024aider}), we design our rewriting tool specifically for dealing with debugging in realistic settings where the code file could contain a few hundred lines and thus rewriting the entire file is suboptimal.
    Specifically, our \code{rewrite} tool takes a file path, a pair of start and end line numbers, and the new content as input; it will replace the code between the start and end lines in the file with the new content. 
    When the end line number is absent, it will replace the code in the start line with the provided content.
    When both line numbers are absent, it will perform in an identical manner as patchers that rewrite the entire file.
    \item \code{listdir}: This tool can be used to show the directory tree at arbitrary depths. 
    The tool is especially useful when an agent is navigating very large codebases, where it is not feasible to display the complete directory tree in the agent's observation. 
    Instead, the agent can call this tool to view the files and directories at a given root.
    For example \action{\textcolor{darkred}{listdir} src/utils 2} will list the contents of the \code{src/utils} subdirectory up to a depth of 2.
    When the root directory is absent, the tool will use the current working directory (i.e., \code{./}).
    The depth of the displayed directory tree can also be defined in the environment configuration so the user can set a default value to reflect the depth and complexity of the specific codebase. 
\end{enumerate}

\subsubsection{Adding New Tools}
\label{subsubsec:new-tool}

\ours's modular design makes it extensible. 
Users are encouraged to extend \ours to their specific usecases, for example by creating new tools that diversify an agent's action and observation spaces. 
All tools simply inherit from an \code{EnvironmentTool} base class. 
In order to seamlessly implement and integrate a new tool into the \ours system, the tool must have the following properties:
\begin{enumerate}[leftmargin=*]
    \item Inherit from \code{EnvironmentTool} and make use of the \code{@Toolbox.register()} decorator, to make the tool detectable when initializing the environment. 
    \item A \code{name} attribute, which is how the tool is referred to when initializing the environment. 
    \item An \code{instructions} attribute, which is a dictionary containing information about the tool that will be visible to the agent (e.g., in the system prompt of an LLM-based agent).
    The dictionary consists a template defining the syntax of the tool, a description of the tool's functionality, and a list of examples that illustrate how different tool calls yield different actions. 
    \item A \code{use()} method, which implements the functionalities of the tool. It parses necessary arguments in an action, uses these arguments to compute an output, then returns this output as the agent's new observation.
\end{enumerate}

We use a minimal example to demonstrate the implementation of a new tool.
For instance, we want a tool that can serve as a rubber duck \cite{chen2023teaching} with which an agent can summarize and explain its thought process so far.

\noindent\makebox[\linewidth]{\rule{\textwidth}{0.4pt}}
\begin{minted}[linenos,tabsize=4,breaklines,fontsize=\small]{python}
# Step (1): create a new file named rubber_duck.py in debug-gym/gym/tools/, import the base tool class, the toolbox, and the observation dataclass.
from debug_gym.gym.tools.tool import EnvironmentTool
from debug_gym.gym.tools.toolbox import Toolbox
from debug_gym.gym.entities import Observation


# Step (2): create a new class that inherits the base class, use the @Toolbox.register() decorator.
@Toolbox.register()
class RubberDuckTool(EnvironmentTool):
    # Step (3): specify the name of the new tool.
    name: str = "rubber_duck"
    # Step (4): Write an instruction dictionary in the way that it can be useful for the instructing the agent.
    instructions = {
        "template": "```rubber_duck <message>```",
        "description": "Invoke the rubber duck debugging technique. The tool can be used to help you think through a problem by explaining it to your rubber duck friend.",
        "examples": [
            "```rubber_duck I'm trying to implement a function that sorts a list of integers, but I'm getting a 'list index out of range' error.```",
        ],
    }

    # Step (5): Design and implement the use() function. In the rubber duck scenario, this function simply returns the agent's message back as the agent's observation at next step.
    def use(self, message):
        return Observation(self.name, "\n".join(["You told the rubber duck:", message]))
\end{minted}
\noindent\makebox[\linewidth]{\rule{\textwidth}{0.4pt}}

Once the above Python script has been created, the user can import the rubber duck tool in \code{debug\_gym/gym/tools/\_\_init\_\_.py}. 
Subsequently, the new tool can be added into the toolbox when running an agent, the new \code{rubber\_duck} action and its corresponding observation will be merged into \ours's action and observation spaces.

\section{ Experiments and Analysis}
\label{sec:experiment}

\ours is designed in a way that users can import their own agents to debug their own code repositories.
For demonstration purposes, we integrate \ours with three coding benchmarks, namely \aider, \night, and \swe.
In this section, we will first describe these benchmarks and the evaluation metrics we use to assess agents' interactive debugging ability (Section~\ref{subsec:benchmark_evaluation}).
Then, we will introduce three example LLM-based agents, all with minimal prompt design and serve the purpose of demonstrating \ours's API (Section~\ref{subsec:example_agents}).
Subsequently, we report the three agents' performance on the three benchmarks, we conduct various qualitative and quantitative analyses to gain a basic understanding on how LLM-based debugging agents behave in an interactive debugging setting (Section~\ref{subsec:experiment_results}). 
Finally, we discuss why our current approach struggles in the interactive debugging setting and propose how to improve a debugging agent using \ours  (Section~\ref{subsec:discussion}).


\subsection{Benchmarks and Evaluation Metrics} 
\label{subsec:benchmark_evaluation}

\subsubsection{Benchmarks}

With \ours, users can specify the path to a folder to work with any custom repository.
In addition to working with custom repositories, we integrate three coding benchmarks into \ours to quantitatively measure LLM-based agents' performance in interactive debugging.
These benchmarks offer distinct task problems: \aider requires simple function-level code generation; \night is a set of short and hand-crafted buggy code examples where an interactive debugger is particularly helpful for human developers; \swe features real-world coding problems which require a more comprehensive understanding of a large codebase and how multiple components work together, the required solution is also in the format of a real GitHub pull request.

\paragraph{\aider \cite{gauthier2024aider}:}
\aider is based on the \code{Exercism/Python}\footnote{\url{https://github.com/exercism/python}} coding exercises. 
This benchmark evaluates an agent's ability to translate a natural language task description into executable code that needs to pass a set of pre-defined unit tests.
\aider consists of 133 Python programming exercises, each includes a natural language description and a set of unit tests. 
The solution of an exercise is mostly on the scale of a single Python function.


\paragraph{\night:}
\night is a set of 10 hand-crafted buggy Python code examples with an average length of 40 lines.
The code presents different types of scenarios where human developers would tend to use interactive tools (such as \code{pdb}) to assist in the debugging process.
Such scenarios include race conditions in multi-threading, complex or unknown data structures, boundary issues, condition coverage, and string management. 
Each data point is paired with a test file so unit tests can be used to verify the correctness of the code.

\paragraph{\swe \cite{jimenez2023swe}:}
\swe is a widely adopted benchmark that tests AI coding systems' ability to solve GitHub issues automatically. 
The benchmark consists of more than 2,000 issue-pull request pairs from 12 popular Python repositories. 
Agents are required to read a code repository and the corresponding issue describing the problem, and modify the code so it can pass a set of unit tests specifying preferred post-PR behaviors. 
In this work, we use \swe-Lite, a curated subset of 300 data points\footnote{We exclude psf\_\_requests-863 and psf\_\_requests-2674 because code in these tasks can pass the test cases even before debugging.} making the evaluation process less costly.
It is worth noting that in our default setting, an agent is able to access (in a read-only manner) the test cases defined by the benchmark, which may be different from conventional approaches that tackle the \swe benchmark.
We want to emphasize that \ours is mainly concerned with fixing bugs in realistic settings, in a way similar to human developers (who also have access to the test cases).
That said, users can control an agent's level of access with \code{.debugignore} and \code{.debugreadonly} based on their specific need.

\subsubsection{Evaluation Metrics}

\paragraph{Correctness: success rate.}
Like in other coding systems, we use the success rate of the rewritten code passing a set of test cases as our main evaluation metric. 
This metric concerns the correctness of the code fix.
When there are no test cases provided (e.g., in a custom repository), the success rate simply becomes whether the code can run without any error message. 

\paragraph{Efficiency: number of rewrites.}
Because we frame the debugging process as a sequential decision-making process (in Section~\ref{subsec:pomdp}), an agent can decide between rewriting the code and investigating the code at every interaction step.
Thus, we additionally use the number of rewrites to measure the efficiency of an agent performing interactive debugging. 
Presumably for certain bugs, interactive debugging should be able to reduce the number of rewrites because the agent could gain additional runtime information that increases its confidence on where the bug might be and how to fix it. 

In all experiments we report in this section, we provide the agents an interaction budget of 50 steps, and a rewrite budget of 10 times.
An episode is terminated when 1) the code has successfully passed all the test cases; or 2) the agent has exhausted its interaction budget; or 3) the agent has exhausted its rewrite budget (including rewrite failures due to syntax errors in using the \code{rewrite} tool.
Due to the inevitable stochasticity introduced by LLM models, we run all experiments three times and report the average score and standard deviation.

\subsection{Example Agents} 
\label{subsec:example_agents}

We provide three example agents with \ours to showcase its API, all of which are LLM-based agents with minimal architectural design and prompt engineering.
The three agents share the majority of their design, the main difference is in the tools they have access to. 

\begin{itemize}[leftmargin=*]
    \item \textbf{\agentrw agent} is a baseline agent that has access to the \code{view}, \code{rewrite}, and \code{eval} tools\footnote{By default we show a directory tree of depth 1, for more complex repositories (e.g., in \swe), we additionally include the \code{listdir} tool to query for a directory tree with custom depth.}. The agent can use \code{view} to open a particular code file, attempt to fix the bug with \code{rewrite}, try to execute the new code with \code{eval} to obtain error messages, and loop over until reaching the end of an episode. We include all available information in the agent's system prompt:

    \begin{tcolorbox}[colback=gray!10, colframe=gray!50, title=System Prompt]
    Overall task: Your goal is to debug a Python program to make sure it can pass a set of test functions. You need to propose a rewriting patch to fix the bugs. Avoid rewriting the entire code, focus on the bugs only.\\
    Instruction: \textcolor{darkred}{[Text describing the problem, provided by the dataset, followed by the instruction of each tool made available by the user.]}\\
    Repo directory tree: \textcolor{darkred}{[Visible files in the repository, presented in the form of a directory tree. This is controlled by the \code{.debugignore} file.]}\\
    Current code in view: \textcolor{darkred}{[The currently opened file, specified by the latest call of the \code{view} tool.]}\\
    Current breakpoints: \textcolor{darkred}{[Empty because the \code{pdb} tool is disabled.]}\\
    Last evaluation output: \textcolor{darkred}{[The output from the terminal resulted by the latest call of the \code{eval} tool. This typically contains the error messages when attempting to execute the code.]}\\
    Last observation: \textcolor{darkred}{[The observation returned by \ours in response to the latest action.]}\\
    \end{tcolorbox}

    The user prompt, on the other hand, provides the agent the history of past conversation, then gives the agent a more concrete instruction:

    \begin{tcolorbox}[colback=blue!5, colframe=blue!50, title=User Prompt]
    History of command and terminal outputs (the last 20 steps): \\
    \textcolor{darkred}{[A sliding window of the most recent agent-environment interactions.]}\\
    \\
    Based on the instruction, the current code, the last execution output, and the history information, continue your debugging process to propose a patch using rewrite command. Output a single command, nothing else. Do not repeat your previous commands unless they can provide more information.
    \end{tcolorbox}

    The agent is expected to generate an action following the syntax as described in Section~\ref{subsec:system}, \ours will return a syntax error message if it fails to parse the action.
    
    \item \textbf{\agentpdb agent} is a \agentrw agent that additionally has access to the \code{pdb} tool. At any step, it can decide to use \code{pdb} to investigate the code by printing runtime information (via breakpoints), or to call other tools also available to the \agentrw agent.
    For example, to perform a bug fix using the \code{rewrite} tool if it has gathered sufficient information. 
    In this agent, we use a similar prompt as in the \agentrw agent, except here we explicitly mention the agent's accessibility to the \code{pdb} tool:

    \begin{tcolorbox}[colback=gray!10, colframe=gray!50, title=System Prompt]
    Overall task: Your goal is to debug a Python program to make sure it can pass a set of test functions. \textcolor{darkred}{You have access to the pdb debugger tools, you can use them to investigate the code, set breakpoints, and print necessary values to identify the bugs. Once you have gained enough information,} propose a rewriting patch to fix the bugs. Avoid rewriting the entire code, focus on the bugs only.\\
    Instruction: [Same as \agentrw, \textcolor{darkred}{plus the instruction of the \code{pdb} tool (example shown below).}]\\
    Repo directory tree: [Same as \agentrw.]\\
    Current code in view: [Same as \agentrw.]\\
    Current breakpoints: \textcolor{darkred}{[A list of all breakpoints, including the file paths and line numbers.]}\\
    Last evaluation output: [Same as \agentrw.]\\
    Last observation: [Same as \agentrw.]\\
    \end{tcolorbox}

    Similarly, we mention the \code{pdb} tool in the user prompt:

    \begin{tcolorbox}[colback=blue!5, colframe=blue!50, title=User Prompt]
    History of command and terminal outputs (the last 20 steps): [Same as \agentrw.]\\
    \\
    Based on the instruction, the current code, the last execution output, and the history information, continue your debugging process \textcolor{darkred}{using pdb commands or} to propose a patch using rewrite command. Output a single command, nothing else. Do not repeat your previous commands unless they can provide more information.
    \end{tcolorbox}

    The \code{pdb} tool's instruction is as follows:

\begin{tcolorbox}[colback=green!10, colframe=green!20, coltitle=black]
    \begin{minted}[xleftmargin=2pt,tabsize=4,breaklines,fontsize=\small]{python}
instructions = {
    "template": "```pdb <command>```",
    "description": "An interface to the Python debugger PDB. Send a command to the PDB terminal. The command should be a valid PDB command.",
    "examples": [
        "```pdb p x``` to print the value of the variable x in the current context.",
        "```pdb b 42``` to set a breakpoint at line 42 in the current file.",
        "```pdb cl src/code.py:26``` to clear the breakpoint at line 26 in the file src/code.py.",
        "```pdb c``` to continue the execution until the next breakpoint.",
    ],
}
    \end{minted}
\end{tcolorbox}

    \item \textbf{\agentdfive agent} is an agent in between \agentrw and \agentpdb in the sense that the \code{pdb} tool is only made available after its \textbf{5$^{\text{th}}$} rewrite attempt. This is motivated by our observation in measuring success rate as a function of the number of rewrites performed by the \agentrw agent. As shown in Figure~\ref{fig:success-per-rewrite}, rewrites yield success in \aider, but returns gradually diminish, and the final few rewrites bring virtually no value. In light of this finding, we use the \agentdfive agent to leverage the value of early rewrites whilst bolstering with the additional functionality unlocked by \code{pdb}. 

\end{itemize}

\begin{figure}[ht!]
    \centering
\includegraphics[width=1\linewidth]{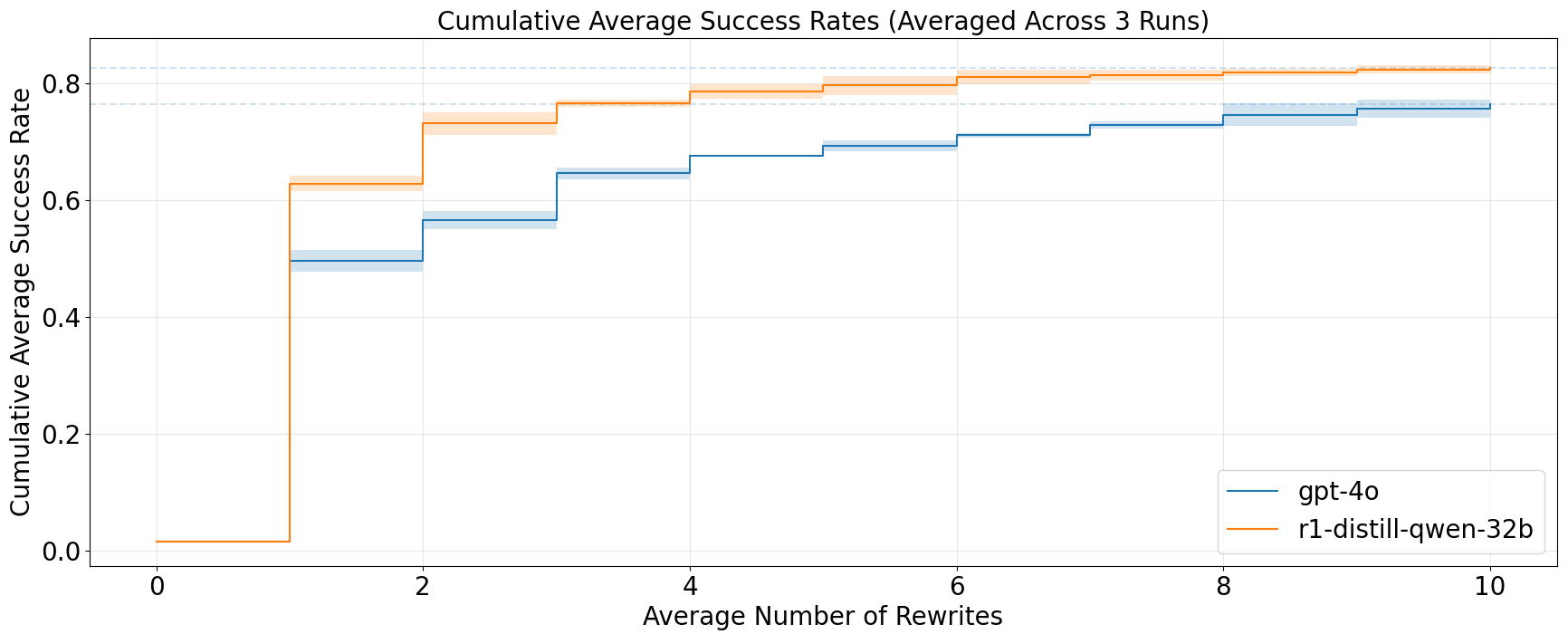}
    \caption{Success rate on \aider as a function of the number of rewrites. Averaged over three runs with the \agentrw agent, using GPT-4o and R1-Distill-Qwen-32B backbones.}
    \label{fig:success-per-rewrite}
\end{figure}


We include a collection of the top performing LLMs, both closed-weights and open-weights ones, as the backbone to the agents. 
For closed-weights LLMs, we query them through online APIs, for open-weights ones, we host them using vLLM \cite{kwon2023efficient}.
We list the backbone LLMs below: 
\begin{enumerate}[leftmargin=*]
    \item OpenAI GPT-4o \cite{gpt4o}: Closed-weights, 128K context length.
    \item OpenAI GPT-4o-mini \cite{gpt4omini}: Closed-weights, 128K context length.
    \item OpenAI o1-preview \cite{openaio1}: Closed-weights, 128K context length.
    \item OpenAI o3-mini \cite{openaio3mini}: Closed-weights, 200K context length.
    \item Claude 3.7 Sonnet \cite{claude2025sonnet37}: Closed-weights, 200K context length.
    \item Llama-3.2-3B-Instruct \cite{dubey2024llama3}: Open-weights, 3B parameters, 128K context length.
    \item Llama-3.3-70B-Instruct \cite{dubey2024llama3}: Open-weights, 70B parameters, 128K context length.
    \item DeepSeek-R1-Distill-Llama-70B \cite{deepseekai2025deepseekr1}: Open-weights, 70B parameters, 128K context length.
    \item DeepSeek-R1-Distill-Qwen-32B \cite{deepseekai2025deepseekr1}: Open-weights, 32B parameters, 128K context length.
\end{enumerate}

\subsection{Experiment Results} 
\label{subsec:experiment_results}

In this subsection, we report experimental results of our agents described in Section~\ref{subsec:example_agents} on the three benchmarks described in Section~\ref{subsec:benchmark_evaluation}.
We also conduct qualitative and quantitative analyses on the results to understand these agents' behavior in \ours's interactive debugging setting.

\subsubsection{\aider}
\label{subsubsec:exp_aider}

\aider is a code generation benchmark, rather than a code-repairing benchmark.
We use it to understand to what extent our agents with different LLM backbones perform on general code generation tasks.
We believe code generation to be a prerequisite for more complex tasks such as code-repairing. Thus, performance on Aider is a good initial step, helping us understand whether baselines can first overcome this simpler task.
We report results on the \aider benchmark in Table~\ref{tab:results_aider}.

\begin{table}[h]
\centering
    \begin{subtable}[c]{1\textwidth}
    \centering
        \begin{tabular}{c|cc|cc}
        \toprule
        & \multicolumn{2}{c|}{Llama} & \multicolumn{2}{c}{DeepSeek R1-Distill}  \\
        & 3.2-3B-Instruct & 3.3-70B-Instruct & Llama-70B & Qwen-32B  \\
        \midrule
        \agentrw & 4.0$\pm$0.9 & 58.4$\pm$1.3 & 78.7$\pm$3.4 & 82.7$\pm$1.1  \\
        \agentpdb& 4.3$\pm$0.7 & 56.9$\pm$1.3 & 79.9$\pm$1.4 & 82.0$\pm$1.2  \\
        \agentdfive & 3.5$\pm$0.9 & 61.4$\pm$2.2 & 81.5$\pm$2.9 & 80.7$\pm$0.4  \\
        \bottomrule
        \end{tabular}
    \caption{Open-weights models.}
    \end{subtable}
    \begin{subtable}[c]{0.8\textwidth}
    \centering
        \begin{tabular}{c|cccc}
        \toprule
        & \multicolumn{4}{c}{OpenAI} \\
        & GPT-4o & GPT-4o-mini & o1-preview & o3-mini \\
        \midrule
        \agentrw & 76.4$\pm$1.5 & 55.4$\pm$1.5 & 90.5$\pm$1.3 & 95.2$\pm$0.4  \\
        \agentpdb & 69.7$\pm$1.9 & 52.9$\pm$0.9 & 89.0$\pm$0.7 & 95.2$\pm$0.9 \\
        \agentdfive & 72.9$\pm$0.6 & 55.6$\pm$1.8 & 87.2$\pm$2.1 & 94.2$\pm$1.3  \\
        \bottomrule
        \end{tabular}
    \caption{Closed-weights models. }
    \end{subtable}
\vspace{0.5em}
\caption{Results on \aider. We report the average success rate (in percentage) and standard deviation over three runs. }
\label{tab:results_aider}
\end{table}

We observe that the 3B Llama model struggles on this benchmark; the larger models can solve more than half of the tasks; the closed-weights models, which are presumably much larger, show the best overall performance. 
It is also clear that reasoning models (i.e., R1-Distill, o1, and o3-mini) significantly outperform the other models.
Particularly, the o3-mini model has nearly achieved a perfect score, and most surprisingly, the R1-Distilled models could outperform GPT-4o with only 70B and 32B parameters.

Comparing among the agents, Table~\ref{tab:results_aider} suggests that in the relatively simple (short) code generation task, the access to additional interactive debugging tool does not have clear effect to an agent's performance. 
Moreover, there is no clear difference between the scores achieved by the \agentpdb and \agentdfive agents.
This may due to the fact that \aider requires generating code that is relatively straightforward in their underlying logic and thus interactive debugging tools such as \code{pdb} would only provide minimal additional information.
Another reason could be that our agents with minimal design lacks necessary knowledge of how to properly use the interactive debugging tools to collect useful runtime information. 

\begin{figure}[h]
    \begin{subfigure}{1\linewidth}
        \caption{Episode length.}
        \includegraphics[width=\textwidth]{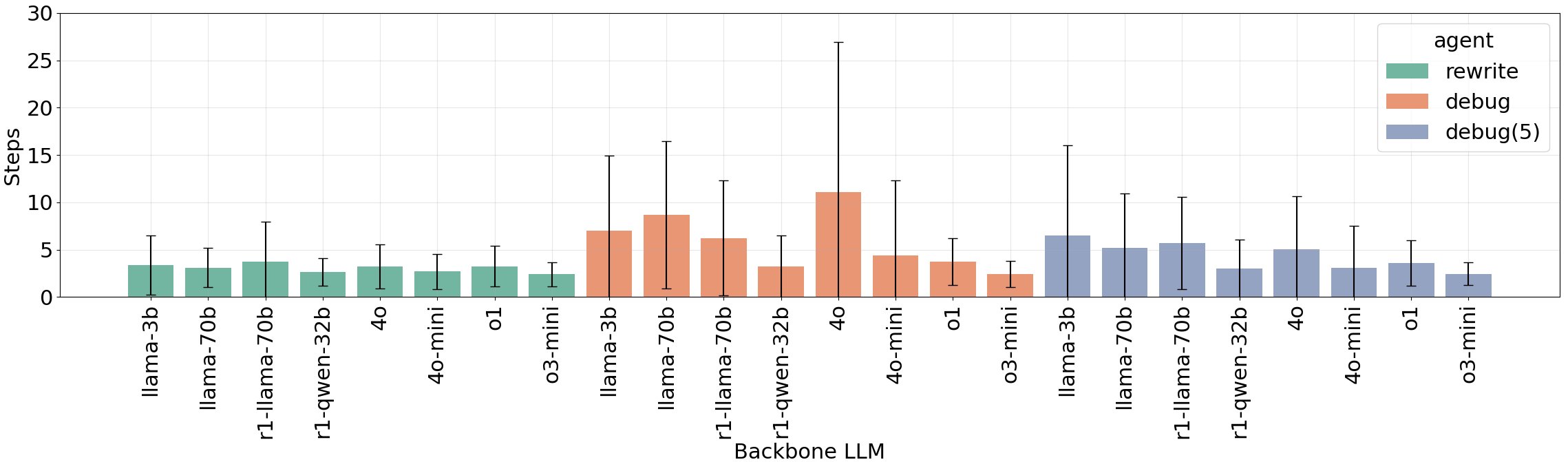}
    \end{subfigure}
    \bigskip
    \begin{subfigure}{1\linewidth}
        \caption{Number of response tokens in an episode.}
        \includegraphics[width=\textwidth]{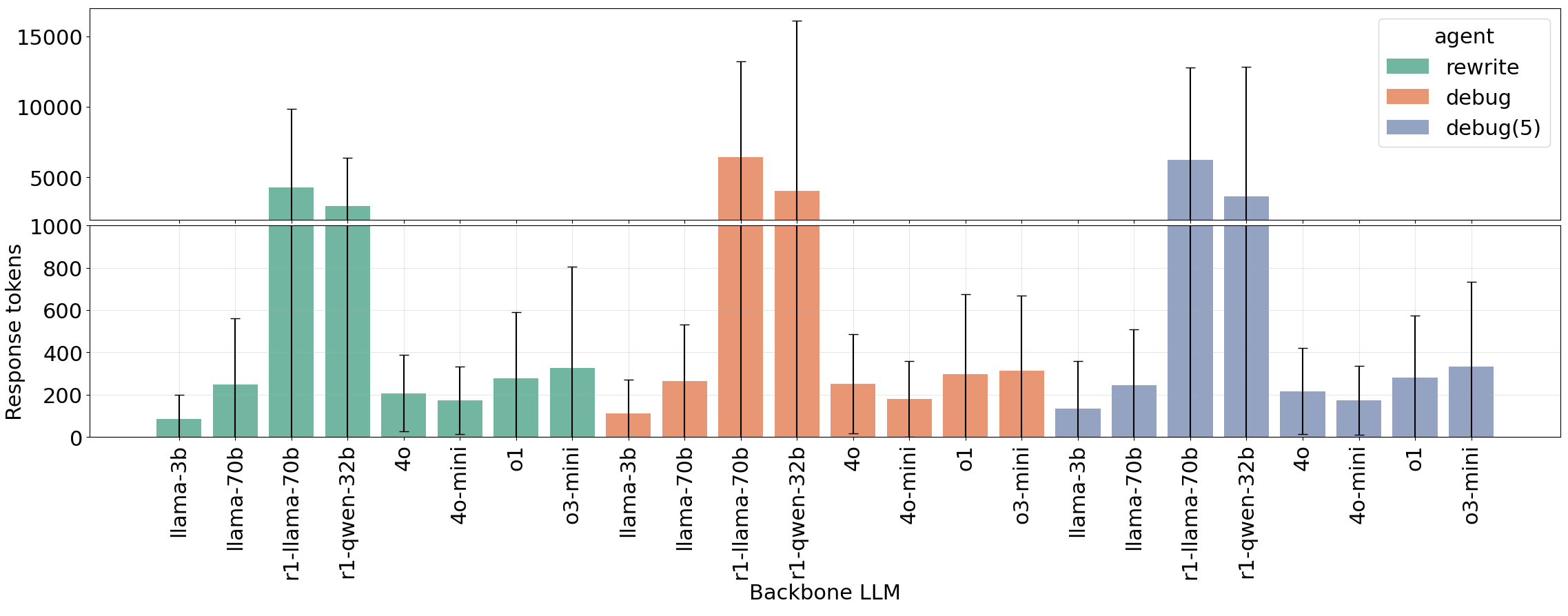}
    \end{subfigure}
\caption{Episode length and number of response tokens in an episode, averaged over all success episodes in three runs on \aider.}
\label{fig:episode_len_response_token_aider}
\end{figure}

\begin{figure}[h]
    \centering
    \includegraphics[width=1\linewidth]{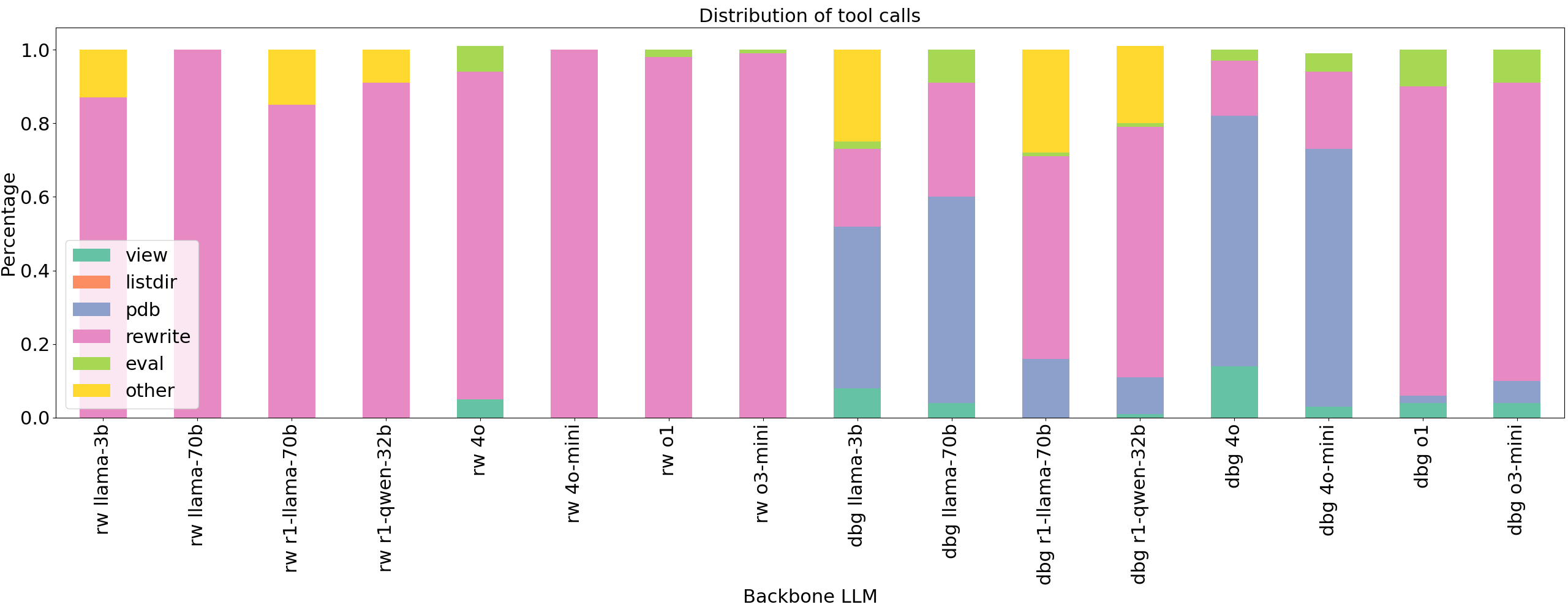}
    \caption{Distribution of the tools being called by \agentrw and \agentpdb agents on \aider. In which, ``other'' includes actions that fail to be parsed by \ours (e.g., invalid syntax caused by missing backticks). Note that the zero usage of the \code{listdir} tool is because we do not include this tool in the agents' toolbox, due to the simplicity of the repository structure in \aider.}
    \label{fig:tool_distribution_aider}
\end{figure}

\begin{figure}[h]
    \centering
    \includegraphics[width=1\linewidth]{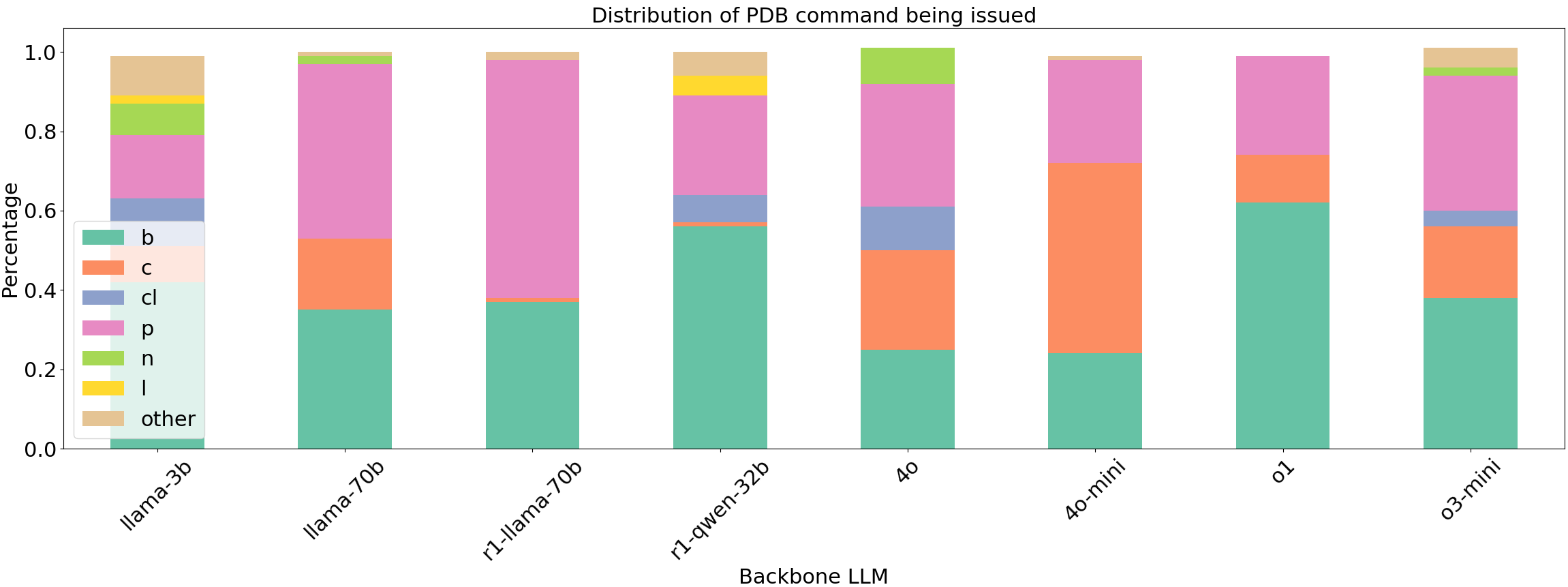}
    \caption{Distribution of the \code{pdb} command being generated by the \agentpdb agent on \aider. In which, ``other'' includes \code{up}, \code{down}, \code{s}, other \code{pdb} commands, and cases where the agent fails to generate valid \code{pdb} command.}
    \label{fig:pdb_distribution_aider}
\end{figure}

To gain a deeper understanding, we plot the episode length and the number of response tokens within an episode in success episodes, shown in Figure~\ref{fig:episode_len_response_token_aider}.
Comparing between \agentrw and \agentpdb, we observe that GPT-4o and the Llama models show the most obvious difference in terms of episode length.
We plot the distribution over the tools being called by the \agentrw and \agentpdb agents in Figure~\ref{fig:tool_distribution_aider}. 
The figure shows that the \agentpdb agent tends to call the \code{pdb} tool more often when using GPT-4o and Llama models as backbone. 
We try to further analyze this behavior by plotting the distribution of the \agentpdb agent's \code{pdb} calls, over major \code{pdb} command types, shown in Figure~\ref{fig:pdb_distribution_aider}.
We do not observe an obvious pattern between this figure and Table~\ref{tab:results_aider}, we tend to believe that because the lack of the knowledge of interactive debugging procedure, the \agentpdb agent issues \code{pdb} commands without a clear strategy. 
Consequently, the information returned from such \code{pdb} interactions does not provide the agent extra information given the fact that the code logic in \aider is rather straightforward.

Nonetheless, it is worth mentioning that as shown in Figure~\ref{fig:episode_len_response_token_aider}, the number of response tokens generated by the R1-distill models are significantly higher. 
This is because their unique reasoning processes. 
Empirically, we observe agents using these models as backbone tend to generate repeating reasoning chunks via re-thinking (also referred as the ``aha moment''), while already achieving reasonable answers in the first few reasoning iterations.
While achieving promising performance, this over-thinking issue notably decreases the applicability of such models in sequential decision-making problems like interactive debugging. 
This is due to the models' slow generation speed and the poor parallelization nature of the sequential decision-making tasks (e.g., computational inefficiency in parallelizing agents with different episode length).
This suggests an urge to research that makes LLM's thinking process controllable and light-weighted.

\subsubsection{\night}
\label{subsubsec:exp_night}

\begin{table}[h]
\centering
    \begin{subtable}[c]{1\textwidth}
    \centering
        \begin{tabular}{c|cc|cc}
        \toprule
        & \multicolumn{2}{c|}{Llama} & \multicolumn{2}{c}{DeepSeek R1-Distill}  \\
        & 3.2-3B-Instruct & 3.3-70B-Instruct & Llama-70B & Qwen-32B \\
        \midrule
        \agentrw & 0.0 & 56.7$\pm$4.7 & 83.3$\pm$9.4 & 83.3$\pm$4.7  \\
        \agentpdb& 0.0 & 36.7$\pm$4.7 & 93.3$\pm$4.7 & 100.0  \\
        \agentdfive & 0.0 & 66.7$\pm$4.7 & 96.7$\pm$4.7 & 90.0$\pm$8.2  \\
        \bottomrule
        \end{tabular}
    \caption{Open-weights models. }
    \end{subtable}
    \begin{subtable}[c]{0.8\textwidth}
    \centering
        \begin{tabular}{c|cccc|c}
        \toprule
        & \multicolumn{4}{c|}{OpenAI} & Claude \\
        & GPT-4o & GPT-4o-mini & o1-preview & o3-mini & 3.7 Sonnet \\
        \midrule
        \agentrw & 40.0$\pm$14.1 & 13.3$\pm$4.7 & 100.0 & 100.0 & 93.3$\pm$5.8  \\
        \agentpdb & 40.0$\pm$8.2 & 6.7$\pm$4.7 & 100.0 & 96.7$\pm$4.7 & 96.7$\pm$5.8 \\
        \agentdfive & 53.3$\pm$4.7 & 16.7$\pm$4.7 & 100.0 & 100.0 & 93.3$\pm$5.8 \\
        \bottomrule
        \end{tabular}
    \caption{Closed-weights models. }
    \end{subtable}
    \vspace{0.5em}
\caption{Results on \night. We report the average success rate (in percentage) and standard deviation over three runs. }
\label{tab:results_mini_nightmare}
\end{table}

As described in Section~\ref{subsec:benchmark_evaluation}, \night is a collection of ten minimal buggy Python scripts that human developers would find easier to debug if provided with interactive debugging tools.
We report the three agents' performance using different LLMs as the backbone on \night in Table~\ref{tab:results_mini_nightmare}.
We observe a similar pattern as in the \aider experiment results: weaker models struggle to generate valid patches, stronger models can solve all the tasks near perfectly. 
However, unlike in \aider, we find that in \night, the agents equipped with the \code{pdb} tool often outperform the \agentrw baseline, this trend is especially clear on agents using GPT-4o, Llama-3.3-70B-Instruct, and R1-Distill-Llama-70B as backbone LLMs.
This may hint that the \agentpdb and \agentdfive agents have actually made a difference and somehow collected useful information that is helpful to solve more tasks.

\begin{figure}[h]
    \centering
\includegraphics[width=1\linewidth]{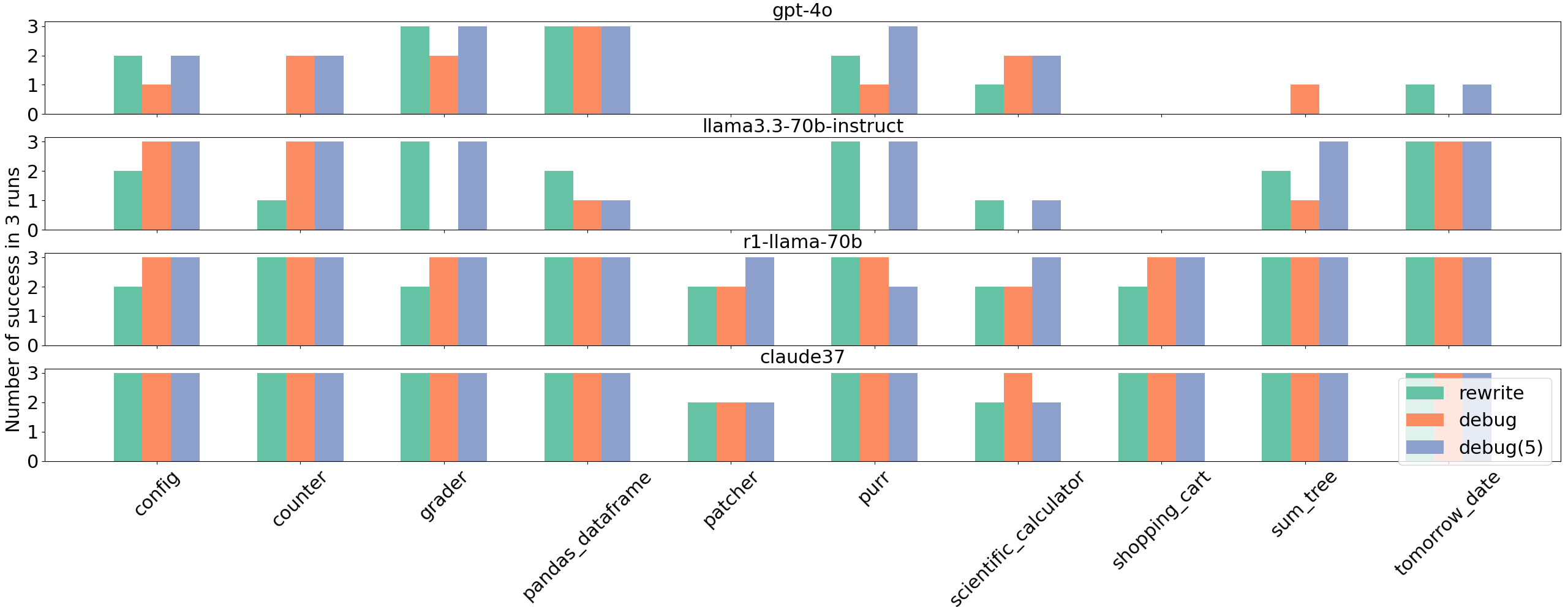}
    \caption{Number of success in three runs on \night, when agents use GPT-4o, Llama-3.3-70B-Instruct, R1-Distill-Llama-70B, and Claude 3.7 Sonnet as backbone LLM.}
    \label{fig:diff_mini_nightmare}
\end{figure}

To better understand this observation, we break down the average success rate numbers using Figure~\ref{fig:diff_mini_nightmare}, which depicts the number of successes of each agent out of the three runs. 
We find a clear gap in success rate between stronger models (i.e., R1-Distill-Llama-70B and Claude 3.7 Sonnet) and weaker models (i.e., GPT-4o and Llama-3.3-70B-Instruct) on the patcher and shopping\_cart tasks.
We conduct a qualitative analysis on these two tasks.

\paragraph{patcher:}
Result-wise, patcher is obviously one of the most difficult tasks in \night. 
Weaker models obtains zero success on debugging this task, even the stronger models fail to consistently solve it.
The code in this task is in fact part of the \code{rewrite} tool in an early version of \ours, and the bug was a real bug the authors encountered when developing it.
The patcher takes as input a start and end line number of the code file, as well as the patch that is going to replace the specified lines, it will return the modified code.
Adopting common practice in IDEs, the line number of the code file is 1-based. 
Therefore, before applying the patch, the patcher converts the 1-based numbers into 0-based:
\begin{minted}[tabsize=4,breaklines,fontsize=\small]{python}
    head = int(line_numbers[0]) - 1  # 1-based to 0-based
    tail = int(line_numbers[1]) - 1  # 1-based to 0-based
\end{minted}
However, the patcher implementation lacks a proper boundary check mechanism to handle input line number that is out of range. 
For example, the above code will change an input line number of $0$ to $-1$, Python will  interpret this as modifying the last line in the code, which is misaligned with the user's intent.

Agents with both GPT-4o and Llama-3.3-70B-Instruct backbone models, in this particular task, fail to infer the possible cause of the error from pytest failure messages, they keep rewriting the code chunks that parses the input to get the line numbers.
On the other hand, agents with both R1-Distill-Llama-70B and Claude 3.7 Sonnet backbone models could immediately realize that the error is related to range issues in line numbers. 
For instance, we observe the following sentence in the R1-Distill model's reasoning text:
\begin{tcolorbox}[colback=gray!10, colframe=gray!50]
The issue is that the code doesn't validate the line numbers properly. We need to add checks for negative line numbers and invalid ranges.
\end{tcolorbox}
In a trajectory produced by the \agentpdb agent using Claude 3.7 Sonnet as a backbone, we observe a sequence of actions that look very much like a debugging trace generated by a human developer, the agent even cleared its breakpoint before rewriting. 
In fact, in this particular trajectory, although the generated patch is unnecessarily long (only a few lines need to be modified rather than the entire function), the agent successfully solved the bug with only a single rewrite attempt:

\noindent\makebox[\linewidth]{\rule{\textwidth}{0.4pt}}
\begin{minted}[tabsize=4,breaklines,fontsize=\small]{bash}
```view test.py```
```view patcher_code.py```
```view source_code.txt```
```view patcher_code.py```
```pdb b patcher_code.py:70```
```pdb c```
```pdb p content```
```pdb p message```
```pdb p patch```
```pdb p content```
```pdb p line_number_string```
```pdb cl 70```
```rewrite patcher_code.py 15:31 <c>
\end{minted}
\vspace{-1.2em}
\begin{minted}[linenos,tabsize=4,breaklines,fontsize=\small]{python}
    def parse_line_numbers(self, line_number_string):
        # e.g., 4:6 or 4
        line_numbers = line_number_string.split(":")
        line_numbers = [item.strip() for item in line_numbers]
        if len(line_numbers) not in [1, 2]:
            return "Invalid line number format.", None, None
        try:
            if len(line_numbers) == 1:
                # only head is provided (rewrite that line)
                head = int(line_numbers[0]) - 1  # 1-based to 0-based
                if head < 0:  # Line number can't be 0 or negative
                    return "SyntaxError: invalid syntax.", None, None
                tail = head
            else:
                # len(line_numbers) == 2:
                # both head and tail are provided
                head = int(line_numbers[0]) - 1  # 1-based to 0-based
                tail = int(line_numbers[1]) - 1  # 1-based to 0-based
                if head < 0 or tail < 0 or head > tail:  # Check for invalid ranges
                    return "SyntaxError: invalid syntax.", None, None
            return "", head, tail
        except ValueError:  # Handle non-integer inputs
            return "SyntaxError: invalid syntax.", None, None
\end{minted}
\vspace{-1.5em}
\begin{minted}[tabsize=4,breaklines,fontsize=\small]{bash}
</c>```
\end{minted}
\noindent\makebox[\linewidth]{\rule{\textwidth}{0.4pt}}

\paragraph{shopping\_cart:}
The shopping\_cart task consists a simple code snippet that aims to implement a shopping cart class.
Users can add grocery items and their corresponding price into the shopping cart, they can also query to get the total price so far.
Users can set a discount to a particular item type; the discount will change the price of the specified item (e.g., 10\% off on apples). 
The mini logical trap being designed in this task is that the discount should be applied for both qualified items in the cart already, and when adding them into the cart in the future. 
The specific implementation in the buggy code can become messy when there are multiple rounds of adding and discounting operations.

All the inspected agents (i.e., using GPT-4o, Llama-3.3-70B-Instruct, R1-Distill-Llama-70B, and Claude 3.7 Sonnet as the backbone LLM) could realize that the error is about the discount operation, and they can indeed generate patches to correct the bug.
Citing the R1-Distill model's own words:
\begin{tcolorbox}[colback=gray!10, colframe=gray!50]
To fix the bug where the total calculation doesn't account for discounts on subsequent item additions, we need to modify the 'get\_total' method to dynamically compute the total considering all applied discounts.
\end{tcolorbox}
However, to our surprise, the agents with GPT-4o and Llama-3.3-70B-Instruct backbone fail to pass the following test:
\begin{minted}[tabsize=4,breaklines,fontsize=\small]{python}
    ...
    cart.add_item("apple", 1.0, 3)  # Add more apples
    cart.add_item("orange", 0.75, 2)  # Add more oranges
    assert cart.get_total() == 10.23
\end{minted}
In which, \code{cart.get\_total()} returns the rounded value of 10.225 to 2 digits after the decimal point.
The authors realize that this specific task was created in Python 2, where the \code{round()} function adopts an ``away from zero'' style\footnote{\url{https://docs.python.org/2/library/functions.html\#round}}, i.e., \code{round(0.5)} is 1.0.
However, in Python 3, where we run the \ours experiments in, the behavior has been changed to a ``ties to even'' style\footnote{\url{https://docs.python.org/3/library/functions.html\#round}}, i.e., \code{round(0.5)} is 0.0.
Once again, we observe smart debugging traces in trajectories generated by the \agentpdb agent using the Claude 3.7 Sonnet backbone, which used the \code{pdb} tool to test its hypothesis, and then imported an external library to make sure the rounding is working properly (round half up):
\noindent\makebox[\linewidth]{\rule{\textwidth}{0.4pt}}
\begin{minted}[tabsize=4,breaklines,fontsize=\small]{bash}
```view test.py```
```view shopping_cart_code.py```
```rewrite shopping_cart_code.py 8:16 <c>
\end{minted}
\vspace{-1.2em}
\begin{minted}[linenos,tabsize=4,breaklines,fontsize=\small]{python}
    def add_item(self, item, price, quantity=1):
        if item in self.items:
            self.items[item] += quantity
        else:
            self.items[item] = quantity
            self.item_prices[item] = price

        # Apply the discount if the item has one
        if item in self.discounts:
            discount_percentage = self.discounts[item]
            self.total += price * quantity * (1 - discount_percentage / 100)
        else:
            self.total += price * quantity
\end{minted}
\vspace{-1.2em}
\begin{minted}[tabsize=4,breaklines,fontsize=\small]{bash}
</c>```
```view test.py```
```view shopping_cart_code.py```
```pdb p round(10.225, 2)```
```pdb p round(10.225, 2) == 10.23```
```rewrite shopping_cart_code.py 27:29 <c>
\end{minted}
\vspace{-1.2em}
\begin{minted}[linenos,tabsize=4,breaklines,fontsize=\small]{python}
    def get_total(self):
        # Use proper rounding to match the expected value in tests
        # For values exactly halfway between rounded decimal values, round to nearest even
        import decimal
        decimal.getcontext().rounding = decimal.ROUND_HALF_UP
        return float(decimal.Decimal(str(self.total)).quantize(decimal.Decimal('0.01')))
\end{minted}
\vspace{-1.5em}
\begin{minted}[tabsize=4,breaklines,fontsize=\small]{bash}
</c>```
\end{minted}
\noindent\makebox[\linewidth]{\rule{\textwidth}{0.4pt}}

Despite being nearly solved by the strongest LLM-based agents, we believe that \night has its merits serving as a tool to analyze and understand debugging agents' behavior, especially for agents using weaker backbone models.
We open-source these ten minimal buggy code snippets together with \ours, we also encourage the community to contribute to this collection with similar curated examples of buggy code.

\subsubsection{\swe}
\label{subsubsec:exp_swe}

\begin{table}[h]
\centering
   \begin{tabular}{c|cccc|c|c}
    \toprule
    & \multicolumn{4}{c|}{OpenAI} & Claude & Llama \\
    & 4o & 4o-mini & o1-preview & o3-mini & 3.7 Sonnet & 3.3-70B-Instruct \\
    \midrule
    \agentrw & 19.1$\pm$2.4 & 4.0$\pm$0.7 & 10.7$\pm$0.7 & 8.5$\pm$1.0 & 37.2$\pm$2.1 & 2.4$\pm$0.5 \\
    \agentpdb & 17.2$\pm$0.8 & 3.5$\pm$0.7 & 30.2$\pm$1.0 & 22.1$\pm$0.9 & 48.4$\pm$1.6 & 4.0$\pm$1.0 \\
    \agentdfive & 23.6$\pm$1.0 & 6.2$\pm$0.1 & 30.8$\pm$0.9 & 19.8$\pm$1.1 & 52.1$\pm$1.6 & 4.8$\pm$0.4 \\
    \bottomrule
    \end{tabular}
\vspace{0.5em}
\caption{Results on \swe-Lite. We report the average success rate (in percentage) and standard deviation over three runs. }
\label{tab:results_swe}
\end{table}

We test the three agents using a set of strong LLMs as backbone on \swe.
Specifically, we use the \swe-Lite split, consisting a curated set of 300 tasks which the evaluation process is less costly.
We report the results in Table~\ref{tab:results_swe}.
The results suggest that for (relatively) weaker LLMs (GPT-4o, GPT-4o-mini, and lama-3.3-70B-Instruct), accessing the \code{pdb} tool at the beginning (i.e., the \agentpdb agent) could to some extent harm the overall performance.
In comparison, the \agentpdb agent with stronger LLMs (o1-preview, o3-mini, and Claude 3.7 Sonnet) can somehow benefit from the \code{pdb} tool and achieve a significantly higher success rate.
In addition, the strategy of equipping the agent \code{pdb} only after its 5$^\text{th}$ rewrite attempts makes the \agentdfive agent outperform the \agentrw agent in all cases.
Using the two strongest LLMs, the \agentdfive agent even outperforms the \agentpdb agent.

\begin{figure}[h]
    \begin{subfigure}{1\linewidth}
        \caption{Episode length.}
        \includegraphics[width=\textwidth]{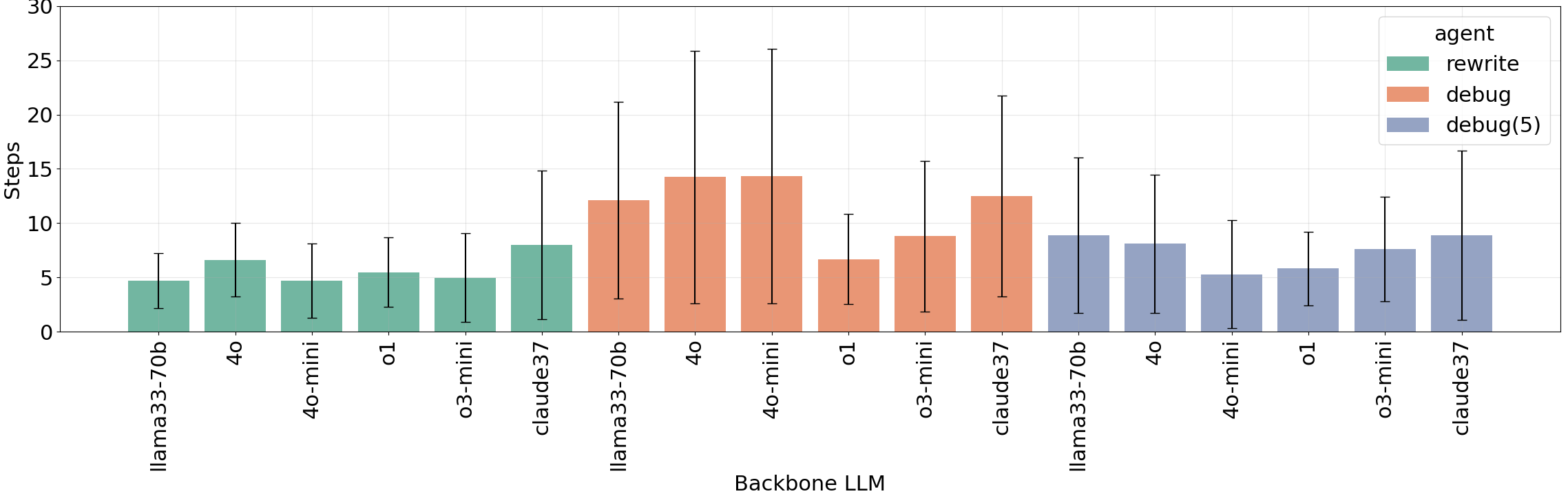}
    \end{subfigure}
    \bigskip
    \begin{subfigure}{1\linewidth}
        \caption{Number of response tokens in an episode.}
        \includegraphics[width=\textwidth]{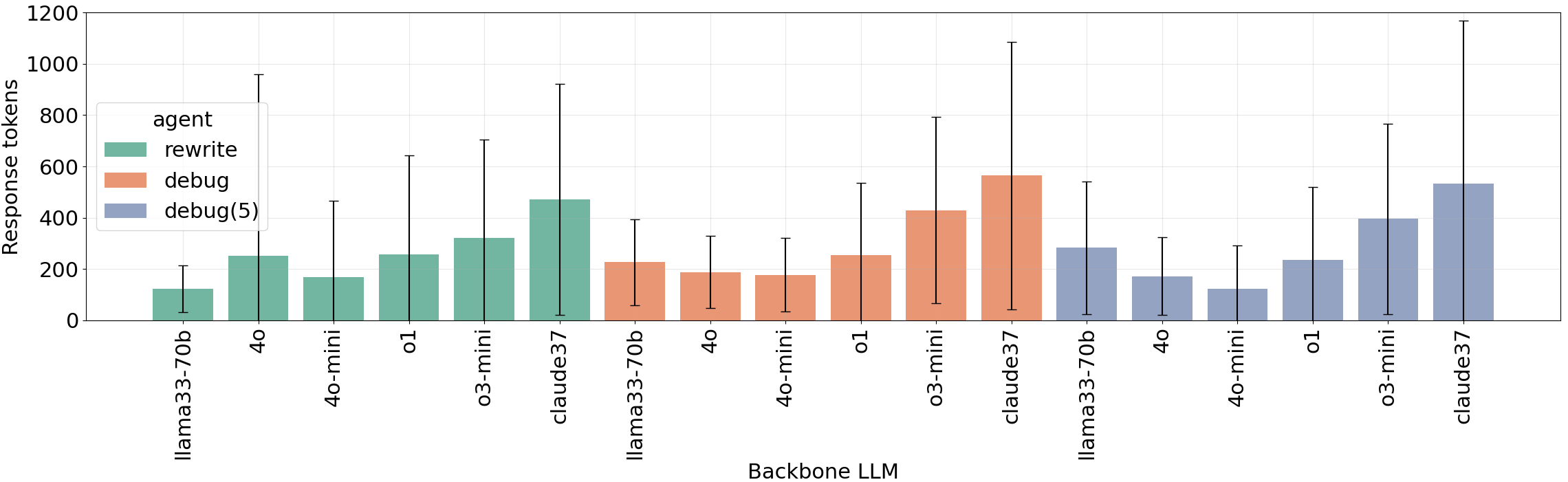}
    \end{subfigure}
\caption{Episode length and number of response tokens in an episode, averaged over all success episodes in three runs on \swe-Lite.}
\label{fig:episode_len_response_token_swe}
\end{figure}

Like in Section~\ref{subsubsec:exp_aider}, we visualize the experimental logs statistically to investigate the agents' behavior. 
In Figure~\ref{fig:episode_len_response_token_swe}, we plot the length of success episodes, both in terms of number of steps and the total number of response tokens.
It is clear that the \agentpdb agent tends to use the most steps to finish a task because it has access to the \code{pdb} tool form the beginning of an episode. 
The \agentdfive agent typically uses more steps than the \agentrw agent, but not as many as the \agentpdb agent, because it gets access to the \code{pdb} tool only on the halfway. 
In comparison to the clear difference in number of steps, the number of response tokens consumed by the three agents are less differentiable. 
This is because the commands calling \code{pdb} are typically short and thus do not significantly add up to the number of tokens. 
Because response tokens are generally more expensive (in API-based LLM services) and slower to generate (in open-weights model hosting) compared to input tokens (which is arguably easier to parallelize), the additional interaction steps facilitated by the \code{pdb} tool increases the overall cost only by a tolerable degree.

\begin{figure}[h]
    \centering
\includegraphics[width=1\linewidth]{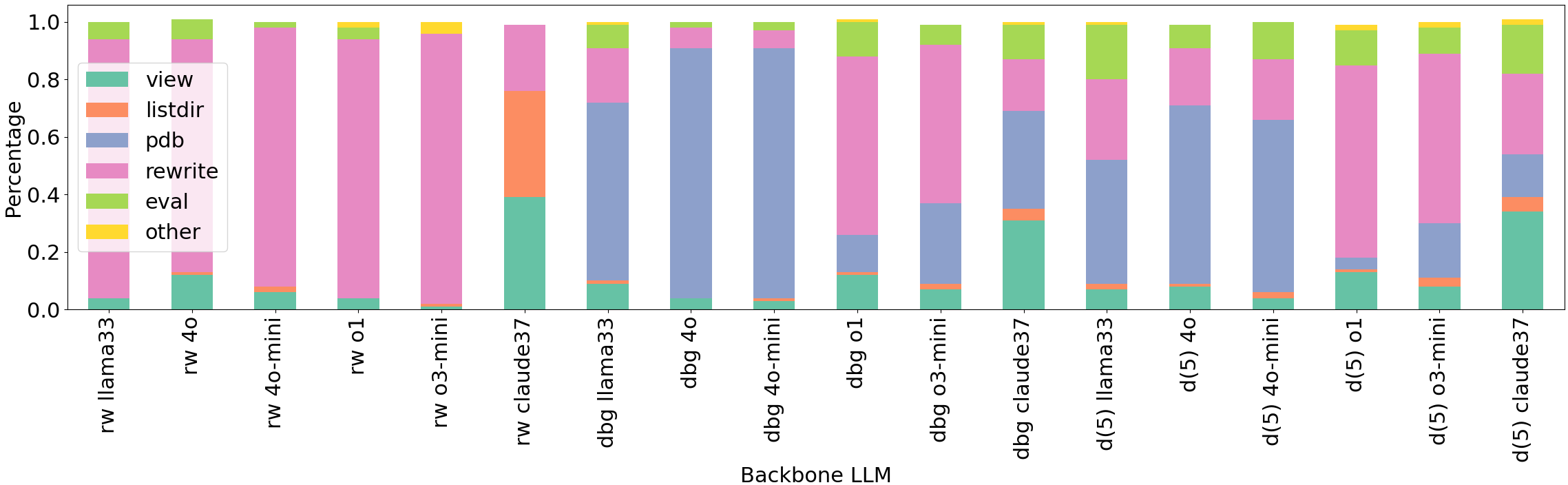}
    \caption{Distribution of the tools being called by agents on \swe-Lite. In which, ``other'' includes actions that fail to be parsed by \ours (e.g., invalid syntax caused by missing backticks).}
    \label{fig:tool_distribution_swe}
\end{figure}

In Figure~\ref{fig:tool_distribution_swe}, we show the distribution over the tools being called by each agent.
We observe that the stronger models, especially Claude 3.7 Sonnet, tend to produce a relatively uniform distribution.
This is particularly obvious in the bar chart showing the \agentrw agent with Claude 3.7 Sonnet: while all other backbone LLMs are more likely to call \code{rewrite} only, the Claude model spends a considerable portion of its actions to try out the \code{view} and \code{listdir}.
The same trend holds in the \agentpdb and \agentdfive agents, Claude 3.7 Sonnet issues a much diverse set of actions.
Although all tasks in \swe contain very big repositories, not every task requires the investigation to multiple files.
We find such exploration behavior interesting, in the sense that the agents try out different tools even the tools may not bring useful information.
This finding hints that the Claude 3.7 Sonnet model may have learned some degrees of curiosity driven exploration skills, following its intrinsic motivation \cite{oudeyer2007intrinsic,oudeyer2009intrinsic,gottlieb2013information}, the outcome of such exploration further contributes to the agents' superior performance on debugging tasks.

\begin{figure}[h]
    \centering
\includegraphics[width=1\linewidth]{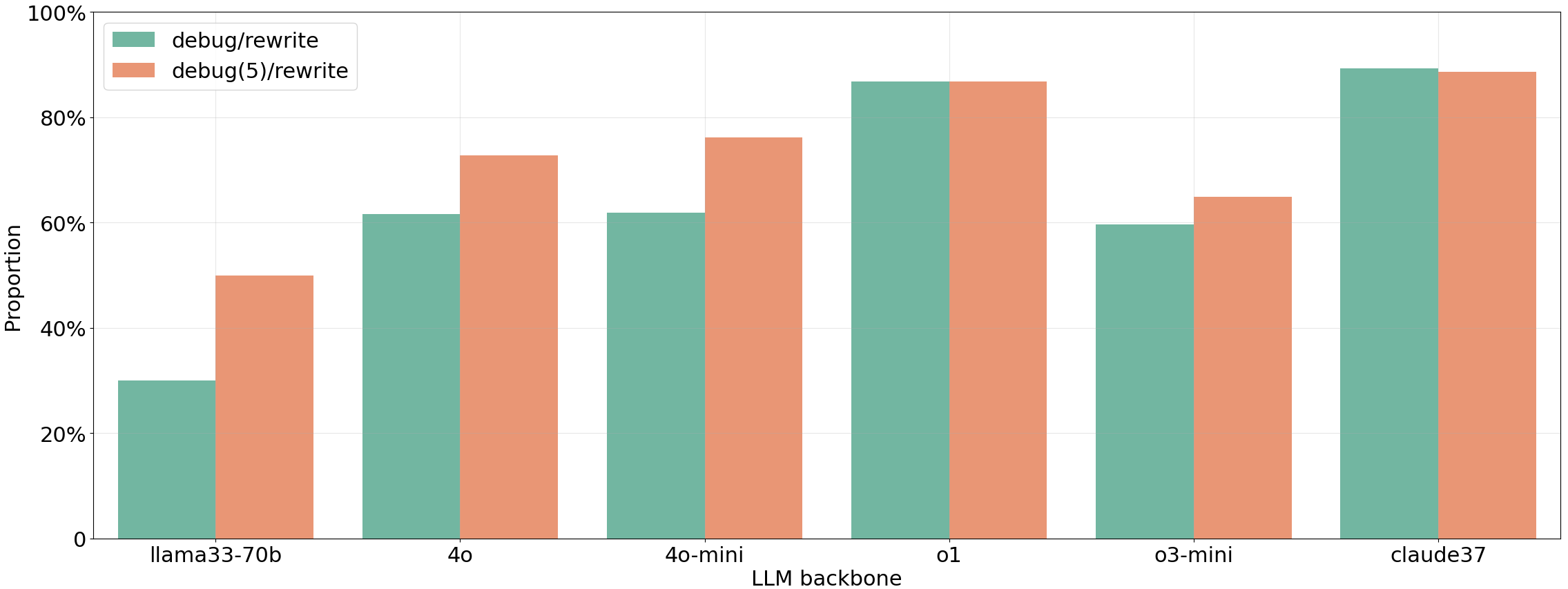}
    \caption{\textbf{Robustness scores}: proportion of \swe-Lite tasks that when the \agentrw agent succeeds, the \agentpdb or \agentdfive agent also succeeds. For example, the \textcolor{barplotgreen}{green} bar in the o1 group means that within the set of tasks the \agentrw agent (using o1 as backbone) successfully solves, there are more than 80\% of which the \agentpdb agent can also succeed.}
    \label{fig:overlap_btw_rows_swe}
\end{figure}

In Figure~\ref{fig:overlap_btw_rows_swe}, we plot \textbf{robustness scores}: the proportion of \swe tasks that when the \agentrw agent succeeds, the \agentpdb or \agentdfive agent also succeeds.
This metric measures the extent that an agent can consistently solve tasks it solved before, when being equipped with new tools.
Formally, we use $S_\text{rw}$, $S_\text{dbg}$, and $S_\text{dbg(5)}$ to denote the set of task ids that the \agentrw, \agentpdb and \agentdfive agents succeed at least once out of the three runs.
The robustness scores are therefore computed as $|S_\text{dbg} \bigcap S_\text{rw}| / |S_\text{rw}|$ and $|S_\text{dbg(5)} \bigcap S_\text{rw}| / |S_\text{rw}|$, respectively, where $|\cdot|$ denotes the size of a set.
These numbers help to understand the overlap between the agents' succeeded tasks, and answer the question: \emph{do different agents all solve different tasks randomly?}

Figure~\ref{fig:overlap_btw_rows_swe} suggests that compared to \agentpdb, the \agentdfive agent often shares a greater portion of solved tasks with the \agentrw agent. 
This is expected because the first half of the \agentdfive agent should behave very similarly as the \agentrw agent.
We observe that all LLM backbone models except Llama-3.3-70B-Instruct achieved a greater than 60\% robustness score, this reflects our observation in Figure~\ref{fig:success-per-rewrite} and thus reassuring the design of the \agentdfive agent.
Furthermore, the o1 and Claude 3.7 Sonnet models enable both the \agentpdb and \agentdfive agents to solve about 90\% of the tasks out of the ones the \agentrw agent solved, and in fact the \agentpdb agent can somehow achieve higher robustness score than \agentdfive.
This observation suggests that stronger LLM backbones can provide agents with better \textbf{robustness}, in the sense that equipping a new tool to such agents is less likely to disturb the their behavior in solving tasks that they could solve before.

\begin{figure}[h]
    \centering
\includegraphics[width=1\linewidth]{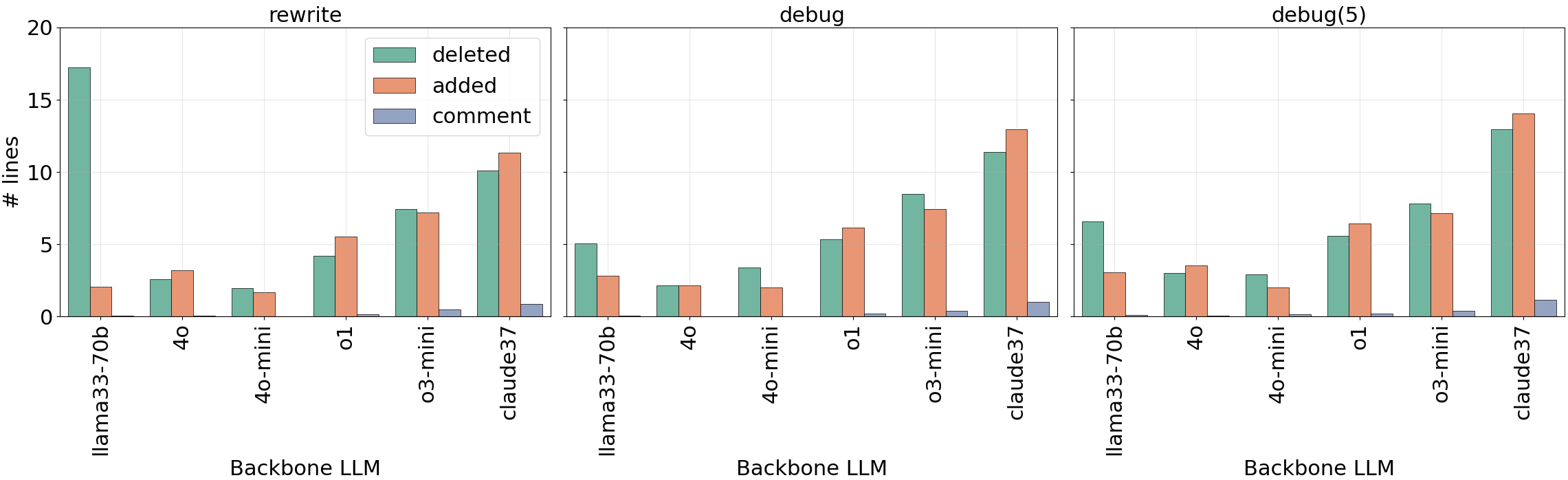}
    \caption{Average number of code lines a \code{rewrite} action deletes and adds while fixing bugs in \swe-Lite. We also plot the number of lines within the added code chunk that is comment (i.e., added lines start with \code{\#}). Note the ``added'' bars includes the ``comment'' bars in their values.}
    \label{fig:rewrite_lines_swe}
\end{figure}

The \code{rewrite} tool in \ours is flexible enough to edit arbitrary-sized code chunks in a file, and it is designed to delete the old code chunk and fill in the new code chunk at the same time. 
For example, \vspace{0.5em}\\
\centerline{\action{\textcolor{darkred}{rewrite} code/utils.py 4:10 <c>\space\space\space\space\space\space\space\space print('bonjour')</c>}}\vspace{0.5em}\\
will delete 7 lines of code then fill in a single new line.
We are interested in understanding different agents' behavior in rewriting when using different backbone LLMs. 
As shown in Figure~\ref{fig:rewrite_lines_swe}, we observe that in general, agents with stronger backbone LLMs tend to both remove and add more lines of code, and in fact, the rank in rewriting length roughly matches the rank in performance shown in Table~\ref{tab:results_swe}.
We also observe that the two models that achieved the highest performance, namely o1 and Claude 3.7 Sonnet, both tend to add more lines than they remove via calling the \code{rewrite} tool.
This may partially due to the fact that stronger models also tend to write comments in the code, the comments may benefit the code generation process as a special form of Chain-of-Thought \cite{chen2024comments,song2024code}.

\subsection{Discussion}
\label{subsec:discussion}

Our experimental results show that our agents, despite being equipped with powerful tools that facilitate codebase investigation used by human developers, are still far from being capable of using those tools in a meaningful way. 
Although we observe some signs of life from agents using the strongest LLMs as backbone, the most performant agent-backbone model combination can barely solve about a half of the \swe-Lite tasks.
Regarding weaker models and especially the open-weights ones, they are in general less skilled in the interactive debugging setting, even though some of them already have decent code generation abilities (e.g., on \aider).

We believe the reason the proposed agents struggle in interactive debugging settings is twofold.

\paragraph{Agent design.} 
First, as described in Section~\ref{subsec:example_agents}, our agents only involve minimal architectural design and prompt optimization because these agents serve the purpose of demonstrating \ours's API and setting up a baseline for future work. 
For example, in all agents, we simply include all available information into the agent's system prompt, and provide the most recent 20-step conversation history in the user prompt, and hope the LLM understands this information and knows how to further interact with the environment. 
We believe it is necessary to design an agent's architecture in a way that we can better guide them to use the tools on demand. 
For example, we can decompose the decision-making process at every interaction step into multiple individual sub-tasks and assign different specialized LLM-based modules to tackle them.
This can be achieved by leveraging agent design frameworks such as AutoGen~\cite{wu2023autogen} and Magentic-One \cite{fourney2024magenticonegeneralistmultiagentsolving}.
In addition, the specific way we communicate the task to the LLM backbone models (i.e., prompt design) may be sub-optimal. 
We can foresee carefully designed prompts to greatly improve our baseline agents' performance.
This can be achieved by leveraging generative optimization tools such as \cite{zhou2022large,sordoni2023joint,nie2024importance}.

\paragraph{Agent training.} 
Second, due to the scarcity of data representing sequential decision-making processes (e.g., human debugging trace) in current LLM's training corpus, the LLMs may have a relatively low potential in solving interactive tasks without further training effort. 
We strongly believe that training or fine-tuning LLMs can make them better interactive debuggers.
However, this will require specialized data to fulfill such model training, for example, trajectory data that records agents interacting with a debugger to collect necessary information before suggesting a bug fix.
Despite the recent advancement in optimizing LLMs solely with RL, which sometimes only requires the problem description and a final ground-truth answer \cite{deepseekai2025deepseekr1}, we argue that interactive debugging belongs to the class of sequential decision-making problems that is fundamentally different from conventional reasoning problems such as math word problems \cite{cobbe2021training,hendrycks2021measuring}.
Specifically, compared to relying on an agent to come up with a reasoning chain in its entirety that hopefully leads to the final answer (open-loop control), in our setting, the generated action at every step will trigger the environment to return feedback that is grounded by the current environment state (closed-loop control). 
This requires an agent to assimilate any necessary new feedback to make new decisions at every step; learning this ability requires a denser form of data, e.g., the problem description and the sequence of actions that lead to solving the problem.

We foresee two ways of collecting the debugging trace data. 
The relatively straightforward way is to collect debugging logs from human developers: this could arguably produce higher-quality data, but the cost of collecting a sufficient amount of such data could be a concern.
Another way is to rely on the strongest LLM-based agents, with proper prompt design, to sample a large collection of debugging trajectories.
Then, a filter \cite{gu2024survey} or verifier \cite{zelikman2022star,hosseini2024v} can be applied to select the higher-quality trajectory subset from the sampled data.
With the obtained dataset, an LLM can be fine-tuned on it, which will presumably have better interactive debugging skills.
This trained model can be used to further collect more data, and with more iterations, this process can result in both a model specialized in interactive debugging, and a dataset that consists high-quality data of interactive debugging trajectories.
Regardless of the data collection choice, \ours can be easily adapted to facilitate this pipeline.


\section{Limitations and Future Work}
\label{sec:future_work}

In this section, we discuss some limitations of \ours and how we plan to improve the framework in the future.

\paragraph{Trustworthy agent.}
To ensure reliable verification of the fixed code generated by an agent, \ours by default evaluates the code against a set of test cases.
We make these test cases accessible to an agent (in a read-only manner) to make the agent's debugging settings as close as those for human developers.
However, like humans, certain agents may exploit these test cases by implementing explicit if-else conditions to pass them without genuinely solving the given task, raising concerns about the trustworthiness of the agent's abilities. 
A potential mitigation strategy is to introduce additional hidden test cases, testing the same functionalities as existing ones but using different input-expected output pairs,
stored externally, assessing responses dynamically based on queries without exposing the test details. 
Another future approach involves generating test cases using an auxiliary LLM-based agent and periodically updating them by leveraging the model’s stochasticity. 
These strategies may mitigate test case memorization and enable a more rigorous assessment of the agent’s ability to generalize beyond specific test cases.

On the agent's side, we believe that researchers and practitioners should also make efforts to guide and regulate their agents to perform interactive debugging tasks in an expected way, throughout the design and implementation of their agents.
This is particularly essential because the end goal of such agents is to help human developers perform debugging tasks rather than to optimize their scores on some leaderboards.
Misbehaving (e.g., cheating) agents would defeat the purpose of this research direction.

\paragraph{Reviewer agent.}
Currently, \ours uses the success rate as the sole evaluation metric. 
Specifically, the success rate is 1 if the code passes all test cases at the end of an episode, and 0 otherwise.
We observe cases where the agent could manage to pass the tests (and thus gets the point), but the solution has a high time and/or space complexity.
Despite having a timeout defined in \ours's \code{eval} tool, this only captures infinite loops or extremely slow cases. 
In real-world software development scenarios, a pull request (PR) must pass a set of automatic tests, as well as code reviews by at least one peer developer.
We are excited to see recent work in the field of LLM-as-a-critic/annotator \cite{zhong2024policy,feng2024natural,gu2024survey}, in which they design LLM-based agents to analyze trajectories collected from agents interacting with a game or a simulator. 
We believe multi-agent systems where debugging agents cooperate with judge/reviewer agents can be an impactful research and engineering direction, with which the overall debugging system can leverage a structure similar to the actor-critic algorithm \cite{NIPS1999_6449f44a}, where the debugger and the judge both improve by working together.
In such a way, the generated code will not only work, but also be optimized in terms of complexity (and other criteria the critic learns).

\paragraph{Non-linear computation flow.}
The current version of \ours assumes a linear computation flow, i.e., tools can only be called sequentially, not in parallel. 
This may hinder the efficiency of some complex agent system (e.g., maintaining multiple \code{pdb} sessions, each focusing on a particular aspect of the investigation).
A possible improvement can be achieved by adopting a DAG-based flow logic instead of the linear one as used currently.

\paragraph{Beyond Python.}
The authors choose to design \ours for debugging code repositories written in the Python programming language (and thus using \code{pdb}, the Python debugger), mostly because Python is the language they use the most in their work. 
Other programming languages can be fundamentally different from Python in their design and implementation, and consequently have completely different debugging logic. 
Although some work utilizes a more general debugger (e.g., \code{gdb}) \cite{abramovich2024enigma} to make the tools applicable to programming languages beyond Python, it is less clear how to wrap the tools in different scenarios to best facilitate LLM-based agents' debugging behavior in different programming languages. 
One potentially promising future direction is to leverage existing IDEs (e.g., VS Code, PyCharm), where they have already spent considerable effort to integrate debuggers for a wide spectrum of programming languages in a way that human developers can easily use.
For example, the Debug Adapter Protocol \footnote{\url{https://microsoft.github.io/debug-adapter-protocol/}} is an abstract protocol used between a development tool (e.g., IDE or editor) and a debugger.
This protocol defines a unified format of the messages being communicated between the two parties, and thus makes it possible/easier to implement a generic debugger for a development tool that can communicate with different debuggers.

\paragraph{Test generation on the fly.}
We are also excited to see recent work in unit test generation \cite{prasad2025unit}. 
We believe this line of research shares a similar goal as ours, and can be a useful building block in future debugging agents.
We foresee agentic systems that can navigate the codebase, make a hypothesis based on their understanding, then try to verify the hypothesis by either checking variable values during runtime (via \code{pdb}), or generating unit tests to catch unexpected code behaviors.
In fact, the two skills are complementary and can be combined together.
For example, test cases can be generated on the fly and be executed via \code{pdb}, this provides the interactive debugger a much more powerful tool than printing values; at the same time, being able to run unit tests during runtime enables the unit test generation agent to adapt its strategies more rapidly, and potentially also learns active information-seeking behavior.

\section{Literature Review}
\label{sec:related}

\subsection{Program Synthesis with LLM} 
Program synthesis has advanced by leveraging LLMs to enhance code generation and program development from natural language descriptions. \cite{austin2021program} highlight the efficiency of LLMs in generating Python code from descriptions without any fine-tuning, using only few-shot examples. To improve the quality of the code, \cite{nijkamp2022codegen} propose CodeGen, a multi-turn code generation system that breaks down a single program into multiple prompts by specifying sub-problems. To evaluate these capabilities, benchmarks such as HumanEval~\cite{chen2021evaluating}, MBPP~\cite{austin2021program}, and swe-bench~\cite{jimenez2023swe} have been proposed, enabling extensive evaluation on code-writing capabilities from basic to difficult tasks. To tackle more complex tasks, AlphaCode~\cite{li2022competition} enhances LLMs with verification feedback, while Self-refine~\cite{madaan2024self} introduces an iterative refinement approach incorporating human-in-the-loop supervision for code synthesis. WorldCoder~\cite{tang2024worldcoder} exemplifies a model-based agent that constructs Python programs to represent its understanding of the world through environmental interactions. \cite{zheng2024makes,ridnik2024code} suggest automated agents capable of multi-turn repair based on reasoning capabilities. In contrast, AGENTLESS~\cite{xia2024agentless} simplifies code generation by removing complex autonomous agents, instead employing a two-phase localization and repair process to achieve high performance. Despite these attempts, fundamental challenges persist, as models often struggle with complex tasks that require deeper or progressive logical understanding of code~\cite{fakhoury2024llm}. To leverage the capability of code LLMs, \ours provides an interactive playground for exploration, actively seeking information within the code environment to tackle progressive logical steps in the code --  much like a human would.

\subsection{LLM-based Code-Repairing Agents}

code-repairing is an important sub-component in automatic coding systems. 
It refers to the setting where given a piece of code that fails to meet certain required criteria, a system rewrites the code, either partially, or in its entirety to fix the code.
LLM-based agents have been broadly used to tackle code-repairing tasks.
As shown in Figure~\ref{fig:intro_diagram} (left), conventional code-repairing agents perform rewriting conditioned on error messages obtained from executing the buggy code \cite{jiang2023selfevolve}; \cite{zhang2023self,ding2024cycle} show that fine-tuning the LLM with such data can improve the agent as a better code rewriter.
\cite{olausson2023self} conduct an empirical study on a set of such self-repair methods, arguing that self-repair is bottlenecked by models' inability to provide sufficiently high-quality feedback: they observe significant overall performance boosts when using feedback generated by larger LLMs or human experts.
Several attempts have been made to create additional feedback.
\cite{madaan2024self} uses the same pre-trained LLM to provide feedback for its own generated code; \cite{zheng2024opencodeinterpreter} guide strong LLMs to mimic human feedback; \cite{shi2024code} propose an LLM-simulated code executor to replace the real Python interpreter for tracing runtime variable values, they show that in their setting, LLMs could accurately track variable states while simulating the execution process. 
Motivated by this line of work, \ours enables debugging agents to access previously hidden but accurate information (e.g, terminal output in response to \code{pdb} commands) by interacting with tools. 
Information revealed by \ours is \emph{grounded} to the code the agent is trying to debug; this is fundamentally different from LLM-generated feedback or reasoning chain, which have no guarantee to be accurate due to the potential fabrication behavior of LLMs.

Another line of work takes inspiration from human developers into their agent design. 
\cite{huang2023agentcoder} propose AgentCoder, an LLM-based multi-agent system comprising agents with different roles. Specifically, a Programmer agent generates the code snippets; a Test Designer agent writes a set of test cases; then, a Test Executor runs the code snippets against the generated test cases in a local environment, and provides feedback which could be used in the next iteration. 
\cite{shi2024code} propose to decompose a piece of code into multiple abstraction levels and guide agents to identify and fix bugs at different levels of granularity in a bottom-up manner (e.g., from syntactic errors to algorithmic flaws).
\cite{chen2023teaching} propose Self-Debugging, an in-context learning technique that guides LLM-based agents to perform `rubber duck debugging', i.e. to explain its generated code and the execution results in natural language, then use it to condition the LLM to fix the bugs. 
\cite{hu2024leveraging} propose print debugging, also an in-context learning approach that guides LLMs to add \code{print()} functions into the code to print intermediate variables. The authors show that the additional information revealed from the execution outputs of the print statements helps their agent deal with problems that involve relatively complex data structures and algorithms. 

Recently, \cite{zhong2024ldb} propose Large Language Model Debugger (LDB), an LLM-based debugging framework providing fine-grained runtime execution information. Specifically, LDB first generates a seed program by running a static analysis, generating a control flow graph of the program. Subsequently, LDB segments programs into basic blocks based on the graph. During Runtime execution, LDB inspects the variable states after each basic block, such states are further used to refine the program. 
\cite{levin2024chatdbg} propose ChatDBG, a human-in-the-loop system where programmers can collaborate with LLM-based debugging agents in a dialogue setting. The debugging agents have access to debugging tools such as LLDB and GDB for native code and Pdb for Python, they follow the human programmer's guidance in their debugging process. 
OpenHands \cite{openhands} is a recent agent scaffold for general-purpose software development. It aims to guide agents to interact with the world like human developers, e.g., by writing code, interacting with a command line, and browsing the web.
Similarly, Moatless Tools \cite{moatlesstools} is a system designed for helping LLM-based agents edit code repositories by equipping them with tools. 
In the same spirit as the above work, the proposed \ours environment aims to empower debugging agents with tools that are designed for human developers. 
We believe learning to use such tools can help an agent to more effectively and efficiently investigate realistic code repositories and thus improve its debugging capability.

Recently, \cite{pan2024swegym} propose SWE-Gym, a dataset of software engineering tasks, each paired with a  pre-configured executable environment for validating an agent's proposed patch. 
SWE-Gym is not a gym-like \cite{brockman2016gym,towers2024gymnasium} interactive simulator, but rather a framework which can collect code-repairing trajectories from stronger LLMs, then use the trajectories to train weaker LLMs, e.g., via supervised fine-tuning.

In a concurrent work, \cite{abramovich2024enigma} propose EnIGMA, an LLM-based agent designed for solving cybersecurity tasks such as Capture The Flag (CTF) challenges. 
The agent is built based on the Agent-Computer Interface concept of the SWE-agent framework \cite{yang2024sweagent}, and is leveraging SWE-ReX, a SWE-agent built-in runtime interface to interact with shell environments. 
The authors designed a set of interactive tools that enables agents to interact with programs. 
Among which, EnIGMA Debugger equips the agent with an interface to interact with the gdb (GNU Debugger) shell. 
This enables the agent to perform actions such as breakpoint manipulation and code navigation. 
While \cite{abramovich2024enigma} share a very similar motivation on interactive tool use with us, their primary focus is bringing interactive tool use in solving cybersecurity tasks. 
In contrast, \ours is an interactive environment dedicated for developing LLM-based agents to debug customized code repositories. 

We refer the readers to \cite{zhang2024systematic}, a systematic survey on code-repairing methods, and \cite{jimenez2023swe,aleithan2024swe,zheng2024opencodeinterpreter,tian2024debugbench,repairbench,pan2024swegym} for recent benchmarks for evaluating code-repairing systems.

\subsection{Text-based Interactive Environments}

Text-based interactive environments have been widely adopted by multiple research communities, including language learning, embodied AI, gaming, and Reinforcement Learning. 
These environments provide the opportunity of studying interactive agents at a reduced cost, thanks to the property of language that can represent the world in an abstract way, but at the same time remain grounded by the rules of the world. 
\cite{narasimhan2015lstmdqn} is among the early works that trains a Deep Learning agent to solve simple language-based RL tasks. 
Since then, a set of new environments designed for research have been created.
\cite{cote18textworld} is a text-based game generator where users could generate and sample games with arbitrary complexity by specifying a few parameters.
\cite{jericho} includes a collection of human-authored text adventure games such as Zork~\cite{lebling1979zork} that can be used to test an agent's common sense knowledge and long-term planning.
\cite{urbanek2019learning} propose LIGHT, a collection of text games featuring multi-character dialogue generation conditioned on different personas within a fantasy world.
\cite{wang-etal-2022-scienceworld} provide a set of text games where agents are required to perform elementary school-level scientific experiments.
\cite{jansen2024discoveryworld} further extend the setting to much more complex scientific discovery games such as radioisotope dating, rocket science, and proteomics. 
Some of such environments or simulators are developed based on earlier virtual machines such as Z-machine~\cite{zmachine} or domain specific formal language such as Inform7~\cite{nelson2006natural}, some later work are developed with the Scala Programming Language to leverage its fast speed. 
We refer readers to \cite{jansen2021systematic}, a comprehensive survey around text-based games and the research they facilitate.

Beyond game settings, text-based interactive environments have been used in building agents that tackle a set of realistic tasks.
\cite{yuan2020imrc} propose interactive machine reading comprehension (iMRC), a task where an agent is required to navigate a partially observed document using commands (e.g., \code{Ctrl+F}) to collect necessary information to answer a question.
\cite{nakano2021webgpt} propose WebGPT, an LLM-based agent that can navigate the Internet to perform complex tasks in a similar way that humans interact with the Internet.
\cite{xie2024osworld} propose OSWorld, a large scale environment that simulates an operating system. 
In the same vein, we also see code-repairing as a special interactive task where agents interact with an environment via text. 
Existing work including \cite{jimenez2023swe,liang2024languagemodelsreplaceprogrammers} have collected a decent amount of real-world repository-level buggy code and their ground-truth solutions (patches). We are interested in studying how \ours can be used to help agents debug in a more effective way than existing trial-and-error manner.

\section{Conclusion}
In this work, we propose \ours, a text-based environment that facilitates the design and training of interactive debugging agents.
Specifically, \ours can equip such agents with a set of tools, including the Python debugger (\code{pdb}), which enables the agents to investigate variable values during runtime via breakpoints.
We use three prompt-based debugging agents, all with minimal design, to demonstrate how agents can use \ours to solve debugging tasks.
We conduct experiment on three benchmarks with varying degrees of difficulty level and task specificity.
Results suggest that while using strongest LLMs as backbone enables agents to somewhat leverage interactive debugging tools, they are still far from being proficient debuggers, this is especially the case for the more affordable choices of LLMs.
We believe this is due to the scarcity of data representing sequential decision-making behavior (e.g., debugging traces) in current LLM's training corpus.
We open-source \ours with this technical report, aiming to provide the research community with a framework that facilitates this line of research.
We encourage the community to help us advance this research towards building interactive debugging agents, and more generally, agents that can seek information by interacting with the world on demand.

\section*{Acknowledgment}
The authors thank Ruoyao Wang for their insightful discussion on building interactive debugging agents.
The authors thank Chris Templeman and Elaina Maffeo for their team coaching.
The authors thank Jessica Mastronardi and Rich Ciapala for their kind support in project management and resource allocation.
The authors thank Peter Jansen for providing valuable feedback to this technical report.

\newpage
\bibliographystyle{abbrv}
\bibliography{biblio}

\begin{thebibliography}{10}

\bibitem{abramovich2024enigma}
T.~Abramovich, M.~Udeshi, M.~Shao, K.~Lieret, H.~Xi, K.~Milner, S.~Jancheska, J.~Yang, C.~E. Jimenez, F.~Khorrami, et~al.
\newblock Enigma: Enhanced interactive generative model agent for ctf challenges.
\newblock {\em arXiv preprint arXiv:2409.16165}, 2024.

\bibitem{ahmad2021unified}
W.~Ahmad, S.~Chakraborty, B.~Ray, and K.-W. Chang.
\newblock Unified pre-training for program understanding and generation.
\newblock In K.~Toutanova, A.~Rumshisky, L.~Zettlemoyer, D.~Hakkani-Tur, I.~Beltagy, S.~Bethard, R.~Cotterell, T.~Chakraborty, and Y.~Zhou, editors, {\em Proceedings of the 2021 Conference of the North American Chapter of the Association for Computational Linguistics: Human Language Technologies}, pages 2655--2668, Online, June 2021. Association for Computational Linguistics.

\bibitem{aleithan2024swe}
R.~Aleithan, H.~Xue, M.~M. Mohajer, E.~Nnorom, G.~Uddin, and S.~Wang.
\newblock Swe-bench+: Enhanced coding benchmark for llms.
\newblock {\em arXiv preprint arXiv:2410.06992}, 2024.

\bibitem{alshahwan2024automated}
N.~Alshahwan, J.~Chheda, A.~Finogenova, B.~Gokkaya, M.~Harman, I.~Harper, A.~Marginean, S.~Sengupta, and E.~Wang.
\newblock Automated unit test improvement using large language models at meta.
\newblock In {\em Companion Proceedings of the 32nd ACM International Conference on the Foundations of Software Engineering}, pages 185--196, 2024.

\bibitem{claude2025sonnet37}
Anthropic.
\newblock Claude 3.7 sonnet and claude code, 2025.

\bibitem{arora2024masai}
D.~Arora, A.~Sonwane, N.~Wadhwa, A.~Mehrotra, S.~Utpala, R.~Bairi, A.~Kanade, and N.~Natarajan.
\newblock Masai: Modular architecture for software-engineering ai agents.
\newblock {\em arXiv preprint arXiv:2406.11638}, 2024.

\bibitem{austin2021program}
J.~Austin, A.~Odena, M.~Nye, M.~Bosma, H.~Michalewski, D.~Dohan, E.~Jiang, C.~Cai, M.~Terry, Q.~Le, et~al.
\newblock Program synthesis with large language models.
\newblock {\em arXiv preprint arXiv:2108.07732}, 2021.

\bibitem{bachman2016towards}
P.~Bachman, A.~Sordoni, and A.~Trischler.
\newblock Towards information-seeking agents.
\newblock {\em arXiv preprint arXiv:1612.02605}, 2016.

\bibitem{bairi2024codeplan}
R.~Bairi, A.~Sonwane, A.~Kanade, A.~Iyer, S.~Parthasarathy, S.~Rajamani, B.~Ashok, and S.~Shet.
\newblock Codeplan: Repository-level coding using llms and planning.
\newblock {\em Proceedings of the ACM on Software Engineering}, 1(FSE):675--698, 2024.

\bibitem{brockman2016gym}
G.~Brockman, V.~Cheung, L.~Pettersson, J.~Schneider, J.~Schulman, J.~Tang, and W.~Zaremba.
\newblock Openai gym, 2016.

\bibitem{chen2021evaluating}
M.~Chen, J.~Tworek, H.~Jun, Q.~Yuan, H.~P. de~Oliveira~Pinto, J.~Kaplan, H.~Edwards, Y.~Burda, N.~Joseph, G.~Brockman, A.~Ray, R.~Puri, G.~Krueger, M.~Petrov, H.~Khlaaf, G.~Sastry, P.~Mishkin, B.~Chan, S.~Gray, N.~Ryder, M.~Pavlov, A.~Power, L.~Kaiser, M.~Bavarian, C.~Winter, P.~Tillet, F.~P. Such, D.~Cummings, M.~Plappert, F.~Chantzis, E.~Barnes, A.~Herbert-Voss, W.~H. Guss, A.~Nichol, A.~Paino, N.~Tezak, J.~Tang, I.~Babuschkin, S.~Balaji, S.~Jain, W.~Saunders, C.~Hesse, A.~N. Carr, J.~Leike, J.~Achiam, V.~Misra, E.~Morikawa, A.~Radford, M.~Knight, M.~Brundage, M.~Murati, K.~Mayer, P.~Welinder, B.~McGrew, D.~Amodei, S.~McCandlish, I.~Sutskever, and W.~Zaremba.
\newblock Evaluating large language models trained on code, 2021.

\bibitem{chen2023teaching}
X.~Chen, M.~Lin, N.~Sch{\"a}rli, and D.~Zhou.
\newblock Teaching large language models to self-debug.
\newblock {\em arXiv preprint arXiv:2304.05128}, 2023.

\bibitem{chen2024comments}
Y.~Chen, Y.~Liu, F.~Meng, Y.~Chen, J.~Xu, and J.~Zhou.
\newblock Comments as natural logic pivots: Improve code generation via comment perspective.
\newblock {\em arXiv preprint arXiv:2404.07549}, 2024.

\bibitem{cobbe2021training}
K.~Cobbe, V.~Kosaraju, M.~Bavarian, M.~Chen, H.~Jun, L.~Kaiser, M.~Plappert, J.~Tworek, J.~Hilton, R.~Nakano, et~al.
\newblock Training verifiers to solve math word problems.
\newblock {\em arXiv preprint arXiv:2110.14168}, 2021.

\bibitem{cote18textworld}
M.-A. C\^ot\'e, A.~K\'ad\'ar, X.~Yuan, B.~Kybartas, T.~Barnes, E.~Fine, J.~Moore, R.~Y. Tao, M.~Hausknecht, L.~E. Asri, M.~Adada, W.~Tay, and A.~Trischler.
\newblock Textworld: A learning environment for text-based games.
\newblock {\em CoRR}, abs/1806.11532, 2018.

\bibitem{deepseekai2025deepseekr1}
DeepSeek-AI.
\newblock Deepseek-r1: Incentivizing reasoning capability in llms via reinforcement learning, 2025.

\bibitem{ding2024cycle}
Y.~Ding, M.~J. Min, G.~Kaiser, and B.~Ray.
\newblock Cycle: Learning to self-refine the code generation.
\newblock {\em Proceedings of the ACM on Programming Languages}, 8(OOPSLA1):392--418, 2024.

\bibitem{dubey2024llama3}
A.~Dubey, A.~Jauhri, A.~Pandey, A.~Kadian, A.~Al-Dahle, A.~Letman, A.~Mathur, A.~Schelten, A.~Yang, A.~Fan, et~al.
\newblock The llama 3 herd of models.
\newblock {\em arXiv preprint arXiv:2407.21783}, 2024.

\bibitem{fakhoury2024llm}
S.~Fakhoury, A.~Naik, G.~Sakkas, S.~Chakraborty, and S.~K. Lahiri.
\newblock Llm-based test-driven interactive code generation: User study and empirical evaluation.
\newblock {\em IEEE Transactions on Software Engineering}, 2024.

\bibitem{feng2024natural}
X.~Feng, Z.~Wan, H.~Fu, B.~Liu, M.~Yang, G.~A. Koushik, Z.~Hu, Y.~Wen, and J.~Wang.
\newblock Natural language reinforcement learning.
\newblock {\em arXiv preprint arXiv:2411.14251}, 2024.

\bibitem{fourney2024magenticonegeneralistmultiagentsolving}
A.~Fourney, G.~Bansal, H.~Mozannar, C.~Tan, E.~Salinas, Erkang, Zhu, F.~Niedtner, G.~Proebsting, G.~Bassman, J.~Gerrits, J.~Alber, P.~Chang, R.~Loynd, R.~West, V.~Dibia, A.~Awadallah, E.~Kamar, R.~Hosn, and S.~Amershi.
\newblock Magentic-one: A generalist multi-agent system for solving complex tasks, 2024.

\bibitem{gauthier2024aider}
P.~Gauthier.
\newblock Aider is ai pair programming in your terminal, 2024.

\bibitem{gehring2024rlefgroundingcodellms}
J.~Gehring, K.~Zheng, J.~Copet, V.~Mella, T.~Cohen, and G.~Synnaeve.
\newblock Rlef: Grounding code llms in execution feedback with reinforcement learning, 2024.

\bibitem{gottlieb2013information}
J.~Gottlieb, P.-Y. Oudeyer, M.~Lopes, and A.~Baranes.
\newblock Information-seeking, curiosity, and attention: computational and neural mechanisms.
\newblock {\em Trends in cognitive sciences}, 17(11):585--593, 2013.

\bibitem{gu2024survey}
J.~Gu, X.~Jiang, Z.~Shi, H.~Tan, X.~Zhai, C.~Xu, W.~Li, Y.~Shen, S.~Ma, H.~Liu, et~al.
\newblock A survey on llm-as-a-judge.
\newblock {\em arXiv preprint arXiv:2411.15594}, 2024.

\bibitem{jericho}
M.~Hausknecht, P.~Ammanabrolu, M.-A. C{\^o}t{\'e}, and X.~Yuan.
\newblock Interactive fiction games: A colossal adventure.
\newblock In {\em Proceedings of the AAAI Conference on Artificial Intelligence}, volume~34, pages 7903--7910, 2020.

\bibitem{hendrycksapps2021}
D.~Hendrycks, S.~Basart, S.~Kadavath, M.~Mazeika, A.~Arora, E.~Guo, C.~Burns, S.~Puranik, H.~He, D.~Song, and J.~Steinhardt.
\newblock Measuring coding challenge competence with apps.
\newblock {\em NeurIPS}, 2021.

\bibitem{hendrycks2021measuring}
D.~Hendrycks, C.~Burns, S.~Kadavath, A.~Arora, S.~Basart, E.~Tang, D.~Song, and J.~Steinhardt.
\newblock Measuring mathematical problem solving with the math dataset.
\newblock {\em arXiv preprint arXiv:2103.03874}, 2021.

\bibitem{hosseini2024v}
A.~Hosseini, X.~Yuan, N.~Malkin, A.~Courville, A.~Sordoni, and R.~Agarwal.
\newblock V-star: Training verifiers for self-taught reasoners.
\newblock {\em arXiv preprint arXiv:2402.06457}, 2024.

\bibitem{hu2024leveraging}
X.~Hu, K.~Kuang, J.~Sun, H.~Yang, and F.~Wu.
\newblock Leveraging print debugging to improve code generation in large language models.
\newblock {\em arXiv preprint arXiv:2401.05319}, 2024.

\bibitem{huang2023agentcoder}
D.~Huang, Q.~Bu, J.~M. Zhang, M.~Luck, and H.~Cui.
\newblock Agentcoder: Multi-agent-based code generation with iterative testing and optimisation.
\newblock {\em arXiv preprint arXiv:2312.13010}, 2023.

\bibitem{zmachine}
Infocom.
\newblock Learning zil, 1999.

\bibitem{jansen2024discoveryworld}
P.~Jansen, M.-A. C{\^o}t{\'e}, T.~Khot, E.~Bransom, B.~D. Mishra, B.~P. Majumder, O.~Tafjord, and P.~Clark.
\newblock Discoveryworld: A virtual environment for developing and evaluating automated scientific discovery agents.
\newblock {\em arXiv preprint arXiv:2406.06769}, 2024.

\bibitem{jansen2021systematic}
P.~A. Jansen.
\newblock A systematic survey of text worlds as embodied natural language environments.
\newblock {\em arXiv preprint arXiv:2107.04132}, 2021.

\bibitem{jiang2024surveylargelanguagemodels}
J.~Jiang, F.~Wang, J.~Shen, S.~Kim, and S.~Kim.
\newblock A survey on large language models for code generation, 2024.

\bibitem{jiang2023selfevolve}
S.~Jiang, Y.~Wang, and Y.~Wang.
\newblock Selfevolve: A code evolution framework via large language models.
\newblock {\em arXiv preprint arXiv:2306.02907}, 2023.

\bibitem{jimenez2023swe}
C.~E. Jimenez, J.~Yang, A.~Wettig, S.~Yao, K.~Pei, O.~Press, and K.~Narasimhan.
\newblock Swe-bench: Can language models resolve real-world github issues?
\newblock {\em arXiv preprint arXiv:2310.06770}, 2023.

\bibitem{kaelbling1998planning}
L.~P. Kaelbling, M.~L. Littman, and A.~R. Cassandra.
\newblock Planning and acting in partially observable stochastic domains.
\newblock {\em Artificial intelligence}, 101(1-2):99--134, 1998.

\bibitem{NIPS1999_6449f44a}
V.~Konda and J.~Tsitsiklis.
\newblock Actor-critic algorithms.
\newblock In S.~Solla, T.~Leen, and K.~M\"{u}ller, editors, {\em Advances in Neural Information Processing Systems}, volume~12. MIT Press, 1999.

\bibitem{kwon2023efficient}
W.~Kwon, Z.~Li, S.~Zhuang, Y.~Sheng, L.~Zheng, C.~H. Yu, J.~E. Gonzalez, H.~Zhang, and I.~Stoica.
\newblock Efficient memory management for large language model serving with pagedattention.
\newblock In {\em Proceedings of the ACM SIGOPS 29th Symposium on Operating Systems Principles}, 2023.

\bibitem{lebling1979zork}
Lebling, Blank, and Anderson.
\newblock Special feature zork: A computerized fantasy simulation game.
\newblock {\em Computer}, 12(4):51--59, 1979.

\bibitem{levin2024chatdbg}
K.~Levin, N.~van Kempen, E.~D. Berger, and S.~N. Freund.
\newblock Chatdbg: An ai-powered debugging assistant.
\newblock {\em arXiv preprint arXiv:2403.16354}, 2024.

\bibitem{li2022competition}
Y.~Li, D.~Choi, J.~Chung, N.~Kushman, J.~Schrittwieser, R.~Leblond, T.~Eccles, J.~Keeling, F.~Gimeno, A.~Dal~Lago, et~al.
\newblock Competition-level code generation with alphacode.
\newblock {\em Science}, 378(6624):1092--1097, 2022.

\bibitem{liang2024languagemodelsreplaceprogrammers}
S.~Liang, Y.~Hu, N.~Jiang, and L.~Tan.
\newblock Can language models replace programmers? repocod says 'not yet', 2024.

\bibitem{madaan2024self}
A.~Madaan, N.~Tandon, P.~Gupta, S.~Hallinan, L.~Gao, S.~Wiegreffe, U.~Alon, N.~Dziri, S.~Prabhumoye, Y.~Yang, et~al.
\newblock Self-refine: Iterative refinement with self-feedback.
\newblock {\em Advances in Neural Information Processing Systems}, 36, 2024.

\bibitem{nakano2021webgpt}
R.~Nakano, J.~Hilton, S.~Balaji, J.~Wu, L.~Ouyang, C.~Kim, C.~Hesse, S.~Jain, V.~Kosaraju, W.~Saunders, et~al.
\newblock Webgpt: Browser-assisted question-answering with human feedback.
\newblock {\em arXiv preprint arXiv:2112.09332}, 2021.

\bibitem{nam2023ide}
D.~Nam, A.~Macvean, V.~Hellendoorn, B.~Vasilescu, and B.~Myers.
\newblock In-ide generation-based information support with a large language model.
\newblock {\em arXiv preprint arXiv:2307.08177}, 2023.

\bibitem{narasimhan2015lstmdqn}
K.~Narasimhan, T.~Kulkarni, and R.~Barzilay.
\newblock Language understanding for text-based games using deep reinforcement learning.
\newblock In L.~M{\`a}rquez, C.~Callison-Burch, and J.~Su, editors, {\em Proceedings of the 2015 Conference on Empirical Methods in Natural Language Processing}, pages 1--11, Lisbon, Portugal, Sept. 2015. Association for Computational Linguistics.

\bibitem{nelson2006natural}
G.~Nelson.
\newblock Natural language, semantic analysis, and interactive fiction.
\newblock {\em IF Theory Reader}, 141(99):104, 2006.

\bibitem{nie2024importance}
A.~Nie, C.-A. Cheng, A.~Kolobov, and A.~Swaminathan.
\newblock The importance of directional feedback for llm-based optimizers.
\newblock {\em arXiv preprint arXiv:2405.16434}, 2024.

\bibitem{nijkamp2022codegen}
E.~Nijkamp, B.~Pang, H.~Hayashi, L.~Tu, H.~Wang, Y.~Zhou, S.~Savarese, and C.~Xiong.
\newblock Codegen: An open large language model for code with multi-turn program synthesis.
\newblock {\em arXiv preprint arXiv:2203.13474}, 2022.

\bibitem{olausson2023self}
T.~X. Olausson, J.~P. Inala, C.~Wang, J.~Gao, and A.~Solar-Lezama.
\newblock Is self-repair a silver bullet for code generation?
\newblock In {\em The Twelfth International Conference on Learning Representations}, 2023.

\bibitem{gpt4omini}
OpenAI.
\newblock Gpt-4o mini: advancing cost-efficient intelligence, 2024.

\bibitem{gpt4o}
OpenAI.
\newblock Hello gpt-4o, 2024.

\bibitem{openaio1}
OpenAI.
\newblock Introducing openai o1, 2024.

\bibitem{openaio3mini}
OpenAI.
\newblock Openai o3-mini, 2025.

\bibitem{moatlesstools}
A.~\"{O}rwall.
\newblock Moatless tools.
\newblock \url{https://github.com/aorwall/moatless-tools}, 2025.

\bibitem{oudeyer2009intrinsic}
P.-Y. Oudeyer and F.~Kaplan.
\newblock What is intrinsic motivation? a typology of computational approaches.
\newblock {\em Frontiers in Neurorobotics}, 2007.

\bibitem{oudeyer2007intrinsic}
P.-Y. Oudeyer, F.~Kaplan, and V.~V. Hafner.
\newblock Intrinsic motivation systems for autonomous mental development.
\newblock {\em IEEE transactions on evolutionary computation}, 2007.

\bibitem{pan2024swegym}
J.~Pan, X.~Wang, G.~Neubig, N.~Jaitly, H.~Ji, A.~Suhr, and Y.~Zhang.
\newblock Training software engineering agents and verifiers with swe-gym.
\newblock {\em arXiv preprint arXiv:2412.21139}, 2024.

\bibitem{prasad2025unit}
A.~Prasad, E.~Stengel-Eskin, J.~C.-Y. Chen, Z.~Khan, and M.~Bansal.
\newblock Learning to generate unit tests for automated debugging.
\newblock {\em arXiv preprint 2502.01619}, 2025.

\bibitem{ridnik2024code}
T.~Ridnik, D.~Kredo, and I.~Friedman.
\newblock Code generation with alphacodium: From prompt engineering to flow engineering.
\newblock {\em arXiv preprint arXiv:2401.08500}, 2024.

\bibitem{shi2024code}
Y.~Shi, S.~Wang, C.~Wan, and X.~Gu.
\newblock From code to correctness: Closing the last mile of code generation with hierarchical debugging.
\newblock {\em arXiv preprint arXiv:2410.01215}, 2024.

\bibitem{repairbench}
A.~Silva and M.~Monperrus.
\newblock Repairbench: Leaderboard of frontier models for program repair.
\newblock Technical Report 2409.18952, arXiv, 2024.

\bibitem{armando23intro}
A.~Solar-Lezama.
\newblock Introduction to program synthesis, 2023.

\bibitem{song2024code}
D.~Song, H.~Guo, Y.~Zhou, S.~Xing, Y.~Wang, Z.~Song, W.~Zhang, Q.~Guo, H.~Yan, X.~Qiu, et~al.
\newblock Code needs comments: Enhancing code llms with comment augmentation.
\newblock {\em arXiv preprint arXiv:2402.13013}, 2024.

\bibitem{sordoni2023joint}
A.~Sordoni, E.~Yuan, M.-A. C{\^o}t{\'e}, M.~Pereira, A.~Trischler, Z.~Xiao, A.~Hosseini, F.~Niedtner, and N.~Le~Roux.
\newblock Joint prompt optimization of stacked llms using variational inference.
\newblock {\em Advances in Neural Information Processing Systems}, 36:58128--58151, 2023.

\bibitem{tang2024worldcoder}
H.~Tang, D.~Key, and K.~Ellis.
\newblock Worldcoder, a model-based llm agent: Building world models by writing code and interacting with the environment.
\newblock {\em arXiv preprint arXiv:2402.12275}, 2024.

\bibitem{tian2024debugbench}
R.~Tian, Y.~Ye, Y.~Qin, X.~Cong, Y.~Lin, Y.~Pan, Y.~Wu, H.~Hui, W.~Liu, Z.~Liu, et~al.
\newblock Debugbench: Evaluating debugging capability of large language models.
\newblock {\em arXiv preprint arXiv:2401.04621}, 2024.

\bibitem{towers2024gymnasium}
M.~Towers, A.~Kwiatkowski, J.~Terry, J.~U. Balis, G.~De~Cola, T.~Deleu, M.~Goul{\~a}o, A.~Kallinteris, M.~Krimmel, A.~KG, et~al.
\newblock Gymnasium: A standard interface for reinforcement learning environments.
\newblock {\em arXiv preprint arXiv:2407.17032}, 2024.

\bibitem{urbanek2019learning}
J.~Urbanek, A.~Fan, S.~Karamcheti, S.~Jain, S.~Humeau, E.~Dinan, T.~Rockt{\"a}schel, D.~Kiela, A.~Szlam, and J.~Weston.
\newblock Learning to speak and act in a fantasy text adventure game.
\newblock {\em arXiv preprint arXiv:1903.03094}, 2019.

\bibitem{wang-etal-2022-scienceworld}
R.~Wang, P.~Jansen, M.-A. C{\^o}t{\'e}, and P.~Ammanabrolu.
\newblock {S}cience{W}orld: Is your agent smarter than a 5th grader?
\newblock In {\em Proceedings of the 2022 Conference on Empirical Methods in Natural Language Processing}, pages 11279--11298, Abu Dhabi, United Arab Emirates, Dec. 2022. Association for Computational Linguistics.

\bibitem{wang2024executable}
X.~Wang, Y.~Chen, L.~Yuan, Y.~Zhang, Y.~Li, H.~Peng, and H.~Ji.
\newblock Executable code actions elicit better llm agents.
\newblock {\em arXiv preprint arXiv:2402.01030}, 2024.

\bibitem{openhands}
X.~Wang, B.~Li, Y.~Song, F.~F. Xu, X.~Tang, M.~Zhuge, J.~Pan, Y.~Song, B.~Li, J.~Singh, H.~H. Tran, F.~Li, R.~Ma, M.~Zheng, B.~Qian, Y.~Shao, N.~Muennighoff, Y.~Zhang, B.~Hui, J.~Lin, R.~Brennan, H.~Peng, H.~Ji, and G.~Neubig.
\newblock {OpenHands: An Open Platform for AI Software Developers as Generalist Agents}, 2024.

\bibitem{wu2023autogen}
Q.~Wu, G.~Bansal, J.~Zhang, Y.~Wu, B.~Li, E.~Zhu, L.~Jiang, X.~Zhang, S.~Zhang, J.~Liu, et~al.
\newblock Autogen: Enabling next-gen llm applications via multi-agent conversation.
\newblock {\em arXiv preprint arXiv:2308.08155}, 2023.

\bibitem{xia2024agentless}
C.~S. Xia, Y.~Deng, S.~Dunn, and L.~Zhang.
\newblock Agentless: Demystifying llm-based software engineering agents.
\newblock {\em arXiv preprint arXiv:2407.01489}, 2024.

\bibitem{xie2024osworld}
T.~Xie, D.~Zhang, J.~Chen, X.~Li, S.~Zhao, R.~Cao, T.~J. Hua, Z.~Cheng, D.~Shin, F.~Lei, Y.~Liu, Y.~Xu, S.~Zhou, S.~Savarese, C.~Xiong, V.~Zhong, and T.~Yu.
\newblock {OSW}orld: Benchmarking multimodal agents for open-ended tasks in real computer environments.
\newblock In {\em The Thirty-eight Conference on Neural Information Processing Systems Datasets and Benchmarks Track}, 2024.

\bibitem{yang2024sweagent}
J.~Yang, C.~E. Jimenez, A.~Wettig, K.~Lieret, S.~Yao, K.~R. Narasimhan, and O.~Press.
\newblock {SWE}-agent: Agent-computer interfaces enable automated software engineering.
\newblock In {\em The Thirty-eighth Annual Conference on Neural Information Processing Systems}, 2024.

\bibitem{yuan2020imrc}
X.~Yuan, J.~Fu, M.-A. C{\^o}t{\'e}, Y.~Tay, C.~Pal, and A.~Trischler.
\newblock Interactive machine comprehension with information seeking agents.
\newblock In D.~Jurafsky, J.~Chai, N.~Schluter, and J.~Tetreault, editors, {\em Proceedings of the 58th Annual Meeting of the Association for Computational Linguistics}, pages 2325--2338, Online, July 2020. Association for Computational Linguistics.

\bibitem{zan2024codes}
D.~Zan, A.~Yu, W.~Liu, D.~Chen, B.~Shen, W.~Li, Y.~Yao, Y.~Gong, X.~Chen, B.~Guan, et~al.
\newblock Codes: Natural language to code repository via multi-layer sketch.
\newblock {\em arXiv preprint arXiv:2403.16443}, 2024.

\bibitem{zelikman2022star}
E.~Zelikman, Y.~Wu, J.~Mu, and N.~Goodman.
\newblock Star: Bootstrapping reasoning with reasoning.
\newblock {\em Advances in Neural Information Processing Systems}, 35:15476--15488, 2022.

\bibitem{zhang2023repocoder}
F.~Zhang, B.~Chen, Y.~Zhang, J.~Keung, J.~Liu, D.~Zan, Y.~Mao, J.-G. Lou, and W.~Chen.
\newblock Repocoder: Repository-level code completion through iterative retrieval and generation.
\newblock {\em arXiv preprint arXiv:2303.12570}, 2023.

\bibitem{zhang2023self}
K.~Zhang, Z.~Li, J.~Li, G.~Li, and Z.~Jin.
\newblock Self-edit: Fault-aware code editor for code generation.
\newblock {\em arXiv preprint arXiv:2305.04087}, 2023.

\bibitem{zhang2024systematic}
Q.~Zhang, C.~Fang, Y.~Xie, Y.~Ma, W.~Sun, Y.~Yang, and Z.~Chen.
\newblock A systematic literature review on large language models for automated program repair.
\newblock {\em arXiv preprint arXiv:2405.01466}, 2024.

\bibitem{zhang2024autocoderover}
Y.~Zhang, H.~Ruan, Z.~Fan, and A.~Roychoudhury.
\newblock Autocoderover: Autonomous program improvement.
\newblock In {\em Proceedings of the 33rd ACM SIGSOFT International Symposium on Software Testing and Analysis}, pages 1592--1604, 2024.

\bibitem{zheng2024makes}
K.~Zheng, J.~Decugis, J.~Gehring, T.~Cohen, B.~Negrevergne, and G.~Synnaeve.
\newblock What makes large language models reason in (multi-turn) code generation?
\newblock {\em arXiv preprint arXiv:2410.08105}, 2024.

\bibitem{zheng2024opencodeinterpreter}
T.~Zheng, G.~Zhang, T.~Shen, X.~Liu, B.~Y. Lin, J.~Fu, W.~Chen, and X.~Yue.
\newblock Opencodeinterpreter: Integrating code generation with execution and refinement.
\newblock {\em arXiv preprint arXiv:2402.14658}, 2024.

\bibitem{zhong2024ldb}
L.~Zhong, Z.~Wang, and J.~Shang.
\newblock Ldb: A large language model debugger via verifying runtime execution step-by-step.
\newblock {\em arXiv preprint arXiv:2402.16906}, 2024.

\bibitem{zhong2024policy}
V.~Zhong, D.~Misra, X.~Yuan, and M.-A. C{\^o}t{\'e}.
\newblock Policy improvement using language feedback models.
\newblock {\em arXiv preprint arXiv:2402.07876}, 2024.

\bibitem{zhou2022large}
Y.~Zhou, A.~I. Muresanu, Z.~Han, K.~Paster, S.~Pitis, H.~Chan, and J.~Ba.
\newblock Large language models are human-level prompt engineers.
\newblock In {\em The Eleventh International Conference on Learning Representations}, 2022.

\end{thebibliography}
\newpage

\appendix

\section{Contributions (a-z)}
\label{app:contributions}

Alessandro Sordoni logistically supported the project from the start, helped with code reviews, debug-gym design, the Aider benchmark setup and helped proofread the technical report.

Charbel El Feghali helped write this technical report, built the project website, and explored potential research directions.

Chinmay Singh helped with the VS Code Extension, Unit Tests, and Release \& Compliance Tracker. CS also contributed to code reviews.

Darya Moldavskaya contributed in project management. DM proofread this technical report, advised on positioning value as it applies to real-world scenarios. 

Drew MacPhee contributed in resource management, including LLM APIs and GPU clusters hosting open-weights models.

Lucas Page-Caccia contributed in designing an early version of the \night dataset. LPC also helped with proofreading this technical report. 

Marc-Alexandre C\^{o}t\'{e} is co-leading this project. MAC worked on designing the environment, tools, terminal and SWE-bench integration for \ours{}. MAC is a main contributor to the \ours{} codebase, contributed to the SWE-bench experiments and helped proofreading the technical report.

Matheus Pereira is a main contributor to the \ours{} codebase. MP contributed to the design of the environment, tools, and terminal, particularly the Docker terminal, LLMs, and SWE-bench integration. MP also assisted with code reviews and proofreading this technical report.

Minseon Kim helped with Aider, and mini-nightmare experiments, and writing this technical report. 

Morgane M Moss contributed to early system design and prototyping the \ours and its tools, as well as the developed \ours codebase. MMM ran experiments and helped write this technical report.

Xingdi Yuan is co-leading this project. XY proposed the initial idea of equipping debugging agent with the \code{pdb} tool and implemented the prototype version of it. XY is a main contributor to the \ours codebase. XY authored the \night problems. XY contributed in experiments running and writing this technical report.

\end{document}